\documentclass{article}

\usepackage{PRIMEarxiv}
\usepackage{amsmath}
\usepackage[utf8]{inputenc} 
\usepackage[T1]{fontenc}    
\usepackage{hyperref}       
\usepackage{url}            
\usepackage{booktabs}       
\usepackage{amsfonts}       
\usepackage{nicefrac}       
\usepackage{microtype}      
\usepackage{lipsum}
\usepackage{fancyhdr}       
\usepackage{graphicx}       
\graphicspath{{images/}} 
\usepackage{algorithm}
\usepackage{algpseudocode}
\usepackage{multirow}
\usepackage[section]{placeins}
\usepackage{tablefootnote}
\usepackage{flafter} 
\usepackage[bottom]{footmisc}
\usepackage[natbibapa]{apacite}
\usepackage{amsmath}

\usepackage{graphicx}

\usepackage{lipsum}

\newcommand\blfootnote[1]{%
  \begingroup
  \renewcommand\thefootnote{}\footnote{#1}%
  \addtocounter{footnote}{-1}%
  \endgroup
}

\algrenewcommand\algorithmicrequire{\textbf{Input:}}
\algrenewcommand\algorithmicensure{\textbf{Output:}}

\title{Policy learning for many outcomes of interest: Combining optimal policy trees with multi-objective Bayesian optimisation
}

\author{
  Patrick Rehill \\
  Centre for Social Research and Methods \\
  Australian National University \\
  Canberra\\
  \texttt{patrick.rehill@anu.edu.au} \\
  \And
  Nicholas Biddle \\
  Centre for Social Research and Methods \\
  Australian National University \\
  Canberra\\
  \texttt{nicholas.biddle@anu.edu.au} \\
}

\begin{document}
\maketitle

\begin{abstract}
Methods for learning optimal policies use causal machine learning models to create human-interpretable rules for making choices around the allocation of different policy interventions. However, in realistic policy-making contexts, decision-makers often care about trade-offs between outcomes, not just single-mindedly maximising utility for one outcome. This paper proposes an approach termed Multi-Objective Policy Learning (MOPoL) which combines optimal decision trees for policy learning with a multi-objective Bayesian optimisation approach to explore the trade-off between multiple outcomes. It does this by building a Pareto frontier of non-dominated models for different hyperparameter settings which govern outcome weighting. The key here is that a low-cost greedy tree  can be an accurate proxy for the very computationally costly optimal tree for the purposes of making decisions which means models can be repeatedly fit to learn a Pareto frontier. The method is applied to a real-world case-study of non-price rationing of anti-malarial medication in Kenya.
\end{abstract}


\section{Introduction}
\blfootnote{\textbf{Statements and Declarations:} No funding was received for conducting this study. The author has no relevant financial or non-financial interests to disclose.}
It is very useful to be able to learn learn human-interpretable rules to make data-driven decisions about the allocation of different treatments to different units. For example, in helping doctors to decide which of several possible therapies will benefit a cancer patient most or in helping governments to decide the eligibility criteria for a job-training program with scarce places. This is what \citet{wald_statistical_1949} originally called learning statistical decision functions and what modern authors have since termed policy learning. The problem of learning good policies for making these decisions is a problem closely related in statistical learning to the problem of causal inference. In fact, policy learning can be seen as implicitly or explicitly learning how causal effects differ across different units and then predicting which rules will best allocate treatments in a population to maximise some measure of value. It therefore straddles both the predictive and explanatory traditions in statistics \citep{breiman_statistical_2001}.

In recent years, the field of policy learning has benefited from breakthroughs in other areas of statistical learning (or more specifically machine learning) which have made it possible to fit much better policy learning models. The first breakthrough was the development of methods to estimate heterogeneous treatment effects in econometrics \citep{wager_estimation_2018,athey_generalized_2019,kunzel_metalearners_2019}.  The second innovation was improvements in the feasibility of fitting optimal decision trees, an NP-hard problem \citep{bertsimas_optimal_2017}.\footnote{For our purposes, NP-hard simply means that the problem scales very poorly as the dataset increases in any dimension. Formally, this means means that the problem is as difficult to solve as the hardest problems in NP, the set of problems only solvable in polynomial time by a non-deterministic Turing machine \citep{fortnow_golden_2013}} These approaches have been combined into policy learning methods that learn policy allocations (essentially as a classification problem classifying units into different policies to maximise overall value) \citep{sverdrup_policytree_2020, bertsimas_optimal_2019}. The optimal tree algorithms here use a substantial amount of computing power compared to their greedy tree alternatives, but they promise a policy that is both human interpretable and provides the global maximum value --- at least on the training data.

Policy learning can offer a lot of value when optimising a single objective, however in policy-making contexts, decision-makers are often interesting in more than one single objective (or at the very least they want to make sure there are not severe trade-offs in other outcomes from optimising on a single objective). For example, the doctor making a decision about treating a cancer patient might be interested both in extending the patient's life by curing or slowing disease progression and also in protecting quality of life by minimising negative side-effects. These goals may be in conflict and so require a decision around trading off one against the other. The government deciding eligibility for a job-training program may want to reduce unemployment as much as possible in aggregate, but they may also want to reduce unemployment for a particular population of concern --- for example a minority group. There may be a trade-off at the margin between prioritising aggregate unemployment and unemployment in this specific group. Obviously in these cases it is not worth simply optimising on one objective, we want a model that allows us to trade-off multiple objectives in some way.

Unfortunately, standard machine learning approaches are only good at minimising loss for a single objective. This has been a problem for other fields in machine learning as well and a number of solutions have been proposed. One proposal is to break down objectives into one 'optimising' and one or more 'satisficing' metrics, here the model is trained to optimise on the former but the model will be rejected if it doesn't meet arbitrary thresholds of performance on the latter \citep{simon_rational_1956}. For example a model to learn job-training program allocation might optimise on reducing unemployment overall but also need to satisfy the metric that on train data it should reduce unemployment for a given minority group at least as much as it does the whole population. Essentially this reduces a multi-objective optimisation problem to a single objective constrained optimisation. However, here we do not get full flexibility to trade-off metrics, we simply set a level for satisficing metrics we wish to achieve and either meet that requirement or not. Another approach is to pick subjective weights \textit{ex-ante} to balance different outcomes in a single combined objective function. However, this decision would be made without understanding the implications of different weightings. There might be trade-offs the decision-maker would have considered subjectively better, however they never knew this option was available because a model was only estimated for one weighting. The ideal solution to this problem should be one that allows a human decision-maker to understand the trade-offs that must be made between the multiple outcomes of interest.

This paper proposes an approach based on the multi-objective Bayesian optimisation (MOBO) literature. Multi-objective optimisers provide a Pareto-optimal or non-dominated set of models across different objectives (i.e. models where it is impossible to improve one objective any further without decreasing another). This means they effectively map out the trade-off between objectives. The proposed approach is termed Multi Objective Policy Learning (MOPoL), it combines the globally optimal \textit{policytree} implemented by \citet{sverdrup_policytree_2020} based on work by \citet{athey_policy_2020} and \citet{zhou_offline_2023} with a MOBO approach that uses evaluation functions (greedy policy trees) that can be fit very quickly to proxy the performance of one that is much more computationally expensive. While it is not feasible to fit many, many optimal trees to map out the Pareto frontier then, MOPoL estimates the frontier so that decision-makers can make a first decision around trading off objectives which they can then refine by fitting a small number of optimal trees to test the performance of the greedy estimate. Finally, an optimal tree can be fit to provide a set of policy rules that fit with the decision-maker's value judgement weighing the trade-off between the objectives.  It is worth noting as well that these objectives need not just be estimated outcomes, but can also be metrics around fairness (e.g. the strength of a fairness regulariser) or transparency (e.g. the number of nodes in the decision tree).

This process involves three different models working together, an evaluation function, a surrogate function and an acquisition function. The evaluation function is simply the function that evaluates the performance of models for a particular set of hyperparameters.\footnote{Hyperparameter is a common machine learning term, though one that is not commonly used in econometrics and so is worth defining explicitly for clarity to a non-machine learning audience. In machine learning a hyperparameter is a parameter that controls the training process rather than one that is optimised in the training process. This means that they need to be tuned by hand or through a search algorithm rather than being explicitly optimised. In a multi-objective context, they can be used to trade-off different objectives making their values not just something we explore to minimise loss, but something we experiment with to derive a set of non-dominated models.} While in many cases of MOBO one would simply use the model that is being tuned to evaluate performance, in the case of MOPoL, optimal policy trees are so computationally challenging that we propose using low-cost proxy functions for evaluation instead. The surrogate function is a model that is used to estimate the performance of the model for unseen hyperparameter settings based on previous iterations in order to identify promising new settings for future iterations. It is iteratively updated based on the results from running an evaluation model. Finally, the acquisition function does the work of searching through the posterior distribution generated by the surrogate model in order to find one or more sets of candidate hyperparameter values to explore on the next iteration.

This process seeks to use this process to learn a policy from data which it will then generalise to a broader population, this of course involves some amount of uncertainty. The concepts of internal validity and external validity are useful in thinking about this uncertainty \citep{neuman_social_2014}. Just because policy learning is not explicitly causal inference does not mean we do not have to think about all the factors that will make a study valid. MOPoL involves estimating doubly robust scores, but these scores still need to reflect the underlying data-generating process in order to implicitly identify a causal effect such that we can learn rules from them (internal validity). In addition, the sample we learn from should be large and representative in order to learn generalisable rules (external validity). The latter of these is much more of a concern in this paper because internal validity ultimate depends on learning good doubly robust scores with machine learning models, a problem that has been discussed at length elsewhere \citep{athey_generalized_2019,zivich_machine_2021,chernozhukov_doubledebiased_2018,westreich_propensity_2010}. The problem of knowing how well a policy will generalise to other populations on two outcomes is one more specific to this paper. It is one we propose resolving in two ways, firstly by quantifying uncertainty around the value that will be produced on multiple outcomes by a given tree by measuring standard error via bootstrapping (hence giving a sense of uncertainty when choosing a weighting hyperparameter value), then on a final policy tree, we hold out a test set to estimate out-of-sample performance for a given set of rules.

This paper’s main contribution is in linking optimal policy learning and multi-objective Bayesian optimisation for the first time. It also provides a new approach to fitting evaluation functions where it is necessary to estimate the performance of an interpretable (although perhaps not particularly powerful) model rather than a powerful black-box model which is the typical context in which MOBO is currently used. Finally it provides an application demonstrating how MOPoL may be used in a real-world policy learning problem using data from a randomised controlled trial on subsidies for anti-malarial treatments in Kenya. Code for this paper can be found at \url{https://github.com/pbrehill/OptimalPolicyMOBO}.

\section{Previous work}
\label{sec:pw}

To the best of our knowledge, there is no existing work that looks to use MOBO to fit optimal policy trees (or optimal decision trees for that matter), however there is a good deal of work on these topics separately which can inform how they might be combined.

\subsection{Learning policy allocation with decision trees}

Policy trees are an adaptation of the Classification and Regression Tree (CART) \citep{breiman_classification_2017} which derive rules to classify the ideal treatment for an individual based on certain conditioning variables. These have been adapted to make policy allocation decisions. Policy learning is a statistical decision-making program with a long history dating back formally at least to \citep{wald_statistical_1949}. The theoretical groundwork for learning policies from observational data was established by Manski who explained the effects that some properties of statistical estimators can have on a learned set of policy rules. This work first covered bias from misidentification \citep{manski_identification_2000} and then the effect of sampling error \citep{manski_statistical_2004}.

Policy learning applications generally place a premium on accuracy due to the high stakes of such decisions and therefore tend to value a globally optimal tree. This is in contrast to other applications where decision trees are used (for example using trees in ensembles) where the dominant approach is fitting greedy trees which optimise at each split, but give no guarantee that they are globally optimal \citep{hu_optimal_2019}. This is because globally optimal trees are extremely computationally expensive and to the extent they are feasible, they depend on a relatively small number of observations, features and low maximum tree-depth \citep{bertsimas_optimal_2017}. However, increases in computing power and some optimisations to algorithms have made such an approach feasible in some cases.

The key motivating factor in the development of policy learning trees has been work on using machine learning approaches to learning heterogeneous treatment effects (HTEs). A widely used method for HTE estimation in causal machine learning is the causal forest \citep{wager_estimation_2018, athey_generalized_2019}, although there is a broader literature on HTE estimation with countless alternative methods \citep{kunzel_metalearners_2019, knaus_machine_2021, carvalho_assessing_2019}. While policy tree does not use an HTE learner per se, it uses machine learning methods to derive doubly robust scores which are also used in R-Learner (of which the causal forest is an implementation) via two nuisance functions --- one estimating outcomes and one estimating treatment propensity. The statistical properties of these scores have led to useful theoretical results around the consistency and robustness of estimates relying on nonparametric outcome and treatment propensity estimates that have proved useful for policy learning as well \citep{athey_policy_2020, zhou_offline_2023}.

Policy trees are tied to ideas around interpretable machine learning, a literature that prioritises machine learning models that can be easily interpreted by a human being (as opposed to more powerful black-box models) \citep{rudin_stop_2019, lipton_mythos_2018}. Decision trees are a widely used model in interpretable machine learning as they create a flow diagram that shows the process for making predictions which is valuable not only for those making decisions based on these models, but also for those who may be affected by these decisions . Policy trees can be understood as a part of this interpretable ML tradition in that they make the prescriptive function of an HTE learner interpretable. For example, it means humans can put these prescriptions into action even without running a new prediction for each new case. It might also be beneficial for the recipient of a treatment or the providers of the treatment to know who is eligible from a clear set of rules. In high-stakes scenarios it may be very important that a human being can understand how these decisions are being made \citep{athey_policy_2020}. However, policy trees as implemented by \citet{sverdrup_policytree_2020} are not strictly an interpretable model because treatment recommendations also depend on the outcomes of black-box nuisance (making them similar to an explainable AI model which fits a secondary model to explain the first \citep{rudin_stop_2019}). This means that while the treatment rules themselves are easy to understand, the process by which the model arrived at those rules is not. This makes these models extremely uninterpretable in terms of what \citet{lipton_mythos_2018} would call algorithmic transparency. They are interpretable in making predictions, but it is difficult to understand how the model was fit to create the decision tree in the first place.

Looking specifically at policy trees, a number of approaches to fitting such a model have been proposed. The simplest approach is to use a single tree for both estimating causal effects and making policy allocation decisions. \citet{kallus_recursive_2017} takes a single, greedy tree approach, \citet{bertsimas_optimal_2019} refines this algorithm and uses a globally optimal implementation. However, this approach limits the power of the causal inference model and may lead to a tree structure that is not ideal for policy allocation as the tree has to serve a dual purpose \citep{amram_optimal_2020}. \citet{athey_policy_2020} and \citet{zhou_offline_2023} proposed using a doubly-robust scores from a black-box HTE model for estimation and then a separate classification model to find policy allocation rules. Doubly robust scores also allow for the derivation of regret bounds on the learned policy equivalent to error bounds on the treatment effect estimates. This approach developed from the approach laid out by \citet{kitagawa_who_2018} which relied on inverse probability weighting to account for sampling bias when learning policies from observational data. However the doubly robust approach is in theory better at reducing sampling bias where we do not have access to oracle propensity scores. This approach was implemented in an R package by \citet{sverdrup_policytree_2020} using a final optimal tree model as suggested by \citet{amram_optimal_2020} and this is the version that this paper employs. This theory was extended by \citet{zhou_offline_2023} to the case of multiple treatments.

\subsection{Multi-objective optimisation}
Multi-objective optimisation (MOO) algorithms provide a way of comparing the performance of different non-dominated models where there are several different outcomes of interest. These could be multiple dependent variables or desiderata not captured in the objective function for example fairness, interpretability, or prediction speed. They produce a set of non-dominated models or a Pareto frontier. This means that in order to improve the performance of a model on one outcome any further, one must reduce performance on another.

There are several different approaches to finding a Pareto frontier, but there are two main sophisticated methods for doing so, evolutionary algorithms and Bayesian optimisation. Evolutionary algorithms evolve a set of solutions by iterating through a process that mimics natural selection \citep{zhou_multiobjective_2011, karl_multi-objective_2022}. They have performed well on many problems and can efficiently produce many Pareto optimal models at once \citep{zhou_multiobjective_2011}. It is possible that evolutionary algorithms could provide an excellent basis for learning a Pareto frontier as they provide a good way to fit multiple near optimal decision trees for a single hyperparameter, however this approach is not readily compatible with existing software tools like \textit{policytree}. This paper instead uses an approach based in Bayesian optimisation as it can be easily adapted to the problem of learning with \textit{policytree}. Bayesian approaches sample hyperparameters from a posterior distribution which represents the likelihood that parameters will provide a non-dominated solution. They evaluate these hyperparameters in a surrogate model to update the posterior. An acquisition function finds a new candidate hyperparameter setting and the process repeats. These surrogate models are traditionally Gaussian Processes, however others model types are possible, for example, the random forest is often used to account for non-continuous hyperparameter spaces \citep{horn_multi-objective_2016, binder_multi-objective_2020}. There are a number of different approaches to MOBO. This paper use a family of approaches based on Expected Hypervolume Improvement (EHVI) which tends to give superior performance and can be parallelised easily compared to alternatives \citep{morales-hernandez_survey_2022, zuhal_comparative_2019, daulton_differentiable_2020}. In particular, it uses the qNEVHI approach which can evaluate q points simultaneously and is more robust to noisy data than other approaches \citep{daulton_differentiable_2020, daulton_parallel_2021}.

\section{Problem set-up}
In policy learning we wish to learn policy allocation rules $\pi$ from data made up of a matrix or vector of outcomes $Y$, a matrix of covariates $X$ and a vector of treatments $\textbf{w} \in Z^+$. For the purposes of this paper we treat the treatment variable as nominal where a single individual only receives one treatment. Per \citet{athey_policy_2020} we learn doubly robust scores $\hat{\Gamma}$  for each outcome and treatment via Augmented Inverse Propensity Weighting (AIPW) \citep{robins_estimation_1994}. As an aside, there is no reason that both the score estimation and policy learning need to use the same covariates; in fact there is nothing stopping the two nuisance functions used to estimate $\hat{\Gamma}$ from using separate sets of variables as well. They could have separate or overlapping variables, however for the sake of simplicity we assume both are simply functions of $X$ (though in the application we explore why this may sometimes not be appropriate). This paper uses the causal forest from the \textit{grf} R package to estimate these scores, though the final stage estimation is actually unnecessary, the easy estimation of doubly robust scores for multiple outcomes makes the package a convenient one to use. Importantly, in order to construct doubly robust scores, it is necessary to properly identify a causal effect. While \citet{athey_policy_2020} focus on the observational case, data would ideally be gathered from an experiment. In this case it is still possible to estimate scores for policy learning \citep{dandl_what_2022}. In this case, Dandl et al. recommend using true treatment propensities in local centering. However, there are advantages to simply locally centering based on estimated scores and this is the approach used in the application \citep{su_when_2023}.

From these scores we then train the policy learner. Per \citet{athey_policy_2020}, the reward values for a given treatment can be estimated directly from $\hat{\Gamma}$. By using scores we hopefully account for the selection bias that may exist in the original observational sample. We can estimate value from these scores in the binary case per \citep{athey_policy_2020} as $$V(\pi,\hat{\Gamma})=\frac{1}{n}\sum^n_{i=1}(2\pi(X_i)-1)\hat{\Gamma}_i.$$In the case of multiple treatments this can be generalised to $$V(\pi,\hat{\Gamma})=\frac{1}{n}\sum^n_{i=1}\langle \pi(X_i),\hat{\Gamma}_i \rangle.$$\citep{zhou_offline_2023} and in the case of multiple outcomes we can express this as a weighted sum of value for different outcomes. $$V(\pi,\hat{\Gamma}, \lambda)=\frac{1}{n}\sum^n_{i=1}\sum_{y \in Y}\lambda_y\langle \pi(X_i),\hat{\Gamma}_{i,y} \rangle, \space \sum_{y \in Y}\lambda_y=1.$$

We can maximise value for a given class of models  $\Pi$ by finding $$\pi^*:=\operatorname{argmax}_\pi(V(\pi,\hat{\Gamma}):\pi \in \Pi).$$This paper restricts the class of models (for reasons of human interpretability and usability) to decision trees of depth $d$ .

For the rest of this paper we talk about value in two different ways. The first is value in the way previously used (as the objective of the learner). This value will be a weighted average of component outcomes governed by some weighting hyper-parameter. Secondly we use value in terms of each given outcome in order to understand the trade-off between outcomes. Here we subscript with the outcome of interest and in effect calculate a value from a model using each outcome's doubly robust scores separately.
$$V_{y}(\pi)=V(\pi(X),\hat{\Gamma}_y).$$Note that this notation for value only applies to outcomes that correspond to a treatment effects estimated from data. Not all objective values need be from the original data though, they could be more general. An objective could be a single scalar like runtime for fitting a model or some kind of measure of model fairness \citep{mehrabi_survey_2019}. In these cases, we can define value more simply as the value on these metrics calculated from the models after fitting and sum these in just as we would with an outcome above. $$V(\pi,\hat{\Gamma}, \lambda)=\frac{1}{n}\sum^n_{i=1}\sum_{y \in Y}\lambda_y \phi_y(\pi(X_i),...), \space \sum_{y \in Y}\lambda_y=1.$$
Here $y$ can be any scalar objective and $\phi_y$ is a function for determining the value on that objective, including $\langle \pi(X_i),\hat{\Gamma}_{i,y} \rangle$. In this paper, we mostly discuss the trade-off on outcomes so we tend to use the more specialised case.

The values of $V(\pi,\hat{\Gamma}, \lambda)$ which are not strictly dominated across all $V_y(\pi)$ make up the Pareto frontier. There is no clear optimisation problem to find this Pareto frontier, rather, this becomes a search problem. We need to map out this frontier by fitting many models representing different trade-offs in an iterative process.  This process should follow the following steps.

\begin{enumerate}
  \item Estimate doubly robust scores for all outcomes $\hat{\Gamma}$ via AIPW.
  \item Pick a hyperparameter value(s) $\lambda$ that can be expected to produce a non-dominated model for a weighting of outcomes (or in the first iteration or first few iterations, pick values at random).
  \item Fit a model with these hyperparameter settings and estimate the value for all objectives.
  \item Iterate through 2 and 3 until a set number of iterations is reached or until it reaches some measure of convergence.
  \item Have a human decision-maker decide which trade-offs are best in their judgement and fit a final model with the chosen hyperparameter settings.
\end{enumerate}

This procedure presents two problems, firstly, it is not obvious what function (called an acquisition function) we should use to decide the most promising hyperparameter values on each iteration. Secondly, an iterative process that requires fitting a large number of models is difficult to do with a computationally expensive fitting algorithm like the optimal policy tree. These questions are explored in Sections 4 and 5 and the solutions proposed are the MOPoL approach which we propose. This involves fitting computationally cheap evaluation models and using the parallel Noisy Expected Hypervolume Improvement (qNEHVI) acquisition function.

Finally, it is worth laying out very clearly what the Pareto frontier is estimating. It is there to estimate the trade-off between outcomes for a policy learner of a certain class (depth-limited decision tree). The particular models themselves are not of interest so much as the curve itself. This means that rather than being interested in the performance of a model on held-out data, we're instead interested in the sampling error of the value estimate and hence standard errors are the best measurement of error at a given point. That is, we care about how much the estimated value generalises to other samples, not the performance of a particular model.

\section{Creating a multi-objective Bayesian optimisation approach for policy learning}
\subsection{Choosing an evaluation function} \label{choosing-eval-fn}
No matter what approach one uses to fit a Pareto frontier, this necessarily means fitting a lot of models with different hyperparameter values. In addition, unlike with some settings for MOBO where models are typically noise-less because there is no need to account for sampling error (for example hyperparameter tuning in a predictive machine learning model), when learning policy rules we have to contend with sampling error that means model evaluations will be noisy estimates of the value. The reason for this is that we are not really interested in the value of the specific model we evaluate, but the value that can be achieved with a learner of given complexity for a given outcome weighting. To contend with this, we will have to estimate a standard error of the value estimate and incorporate it into the surrogate function somehow. Unfortunately as we lack a central limit theorem on value estimates from policy trees we must rely on non-parametric bootstrapping to estimate standard errors. This means fitting each model many times just to trial a single hyperparameter setting. With many models this would be an annoyance, but with optimal trees the analysis is not just annoying, for medium-sized datasets it is infeasible.

The optimal policy tree is a very computationally expensive algorithm. On realistically sized datasets it can be many orders of magnitude slower than greedy or even hybrid trees (i.e. trees that look ahead a certain number of splits when making a split decision but don't look across the whole tree). Per \citet{sverdrup_policytree_2020}, the runtime of the optimal tree is $O(p^kn^k(\operatorname{log}n+ d)+pn\operatorname{log}n$) where $p$ is the number of features, n the number of observations, $d$ the number of treatments, and $k \geq 1$ the tree depth. In order to fit a Pareto frontier (with bootstrapped standard errors) on a dataset of any significant size then, we need to find a way to approximate the performance of the optimal tree more quickly (importantly, we just need to approximate the value generated, the actual rules generated are not important because the final optimal model will find a better set of rules). This paper proposes using a separate evaluation model for this purpose. This model that is easier to fit than the target model, but yields a good-enough approximation of model performance to inform hyperparameter tuning (such as the priority put on different outcomes). This is something that can be validated by examining the shape of the Pareto frontier relative to a handful of optimal models.

While the bias from having a greedy model should be manageable, it is still important to have a model that accurately captures the complexity --- in terms of final model complexity --- of a decision tree. We cannot for example rely on computationally cheaper algorithms that will produce more complex final models (ala $PolicyForest$ \citep{oprescu_econml_2019}) because the whole point of the exercise is to learn a human-understandable set of rules. The policy forest would probably perform much better than the policy tree but at the cost of not actually producing an implementable policy. One way of doing this is to fit a greedy tree, or a hybrid tree where the final model is of like complexity, but the performance of the optimal tree on training data is an upper bound on value. The implication of this bound is that even without knowing the optimal model for a given outcome weighting $\pi_O(X|\lambda)$ and the greedy model that seeks to approximate it $\pi_G(X|\lambda, \theta^*)$ one can state that across hyperparameter settings $\lambda \in \Lambda$

$$V(\pi_O(X, \vert \lambda ), \hat{\Gamma}) \approx V(\pi_G(X, \vert \lambda , \theta^{*}), \hat{\Gamma}),$$ $$\theta^{*} := \operatorname{argmin}_{\theta}[V(\pi_O(X \vert \lambda ), \hat{\Gamma}) - V(\pi_G(X\vert \lambda , \theta), \hat{\Gamma})].$$

This is because we can assume that by the definition of optimal and greedy trees, $V(\pi_O(X\vert \lambda ), \hat{\Gamma})) \geq V(\pi_G(X\vert \lambda , \theta^{*}), \hat{\Gamma} )$, and as so the optimisation has the useful property that the greater $V(\pi_G(X \vert \lambda, \theta^{*}), \hat{\Gamma})$ is, the better $\pi_G$ approximates $\pi_O$ given that both models are simply trying to maximise the value of partitioning the data with a decision tree of a given depth (and the same value for all other tuning parameters as well).

This means we can approximate $\pi_O$ without actually having to fit it. The outcome estimates from the greedy frontier will be biased downwards, but a biased frontier can still be useful because what we are really interested in is the marginal trade-off at a given weighting value. Besides this it is always possible to fit a series of optimal trees on the same hyperparameter values to observe 'ground-truth' tree values of this bias. A researcher might use MOPoL to fit a Pareto frontier (which involves bootstrapping) before double-checking the frontier by trying the same hyperparameter settings with optimal trees as this would not require bootstrapping.

There are many possible algorithms for fitting an evaluation model under these constraints, however this paper considers two – the greedy policy tree and the hybrid policy tree which optimises across the current step and the next step after that at each split.\footnote{Other methods for finding closer-to-optimal decision trees such as those that extract rules from random forests like \citet{domingos_knowledge_1997} or \citet{liu_learning_2014} or evolutionary decision tree algorithms such as \citet{grubinger_evtree_2014} may also be useful, although implementing policy learning in these methods is beyond the scope of this paper so they are not considered.} Due to the different computational cost of these approaches as well as the need to iterate through these models for bootstrapping standard errors and fitting many different points in the hyperparameter-space $\Lambda$, the defining characteristic on which this decision is made is likely to be runtime. Specifically, the decision will be whether the longer runtime of the hybrid tree is acceptable given its higher accuracy. As such, the choice of the ideal evaluation model will depend in large part on the size of the dataset (both in number of observations and number of features), the number of objectives, and the computational budget for fitting it.

It is worth noting as well that given the fact optimal trees scale so poorly as a dataset increases, an optimal tree is likely not the best approach even for fitting a single, final model for many problems with large datasets. It is likely that in many applications the sampling error will be larger than the error from using a greedy algorithm. This means that if the sample size grows large enough that it becomes infeasible to fit an optimal tree, it would be better to simply use a greedy or hybrid tree rather than trimming the sample. This is a trade-off that has been little discussed in the policy learning literature, but is likely to be an important consideration in real-world applications.

\subsection{Acquisition and surrogate functions}
The MOPoL algorithm adapts standard approaches to MOBO to the specific case of learning an optimal policy tree. While MOPoL could use many different approaches to MOBO (e.g. ParEGO, EEI), qNEHVI was chosen due to its strong performance in comparisons with other approaches, suitability for noisy data \citep{daulton_parallel_2021}, ability to acquire several new candidate hyperparameter settings at once for parallel processing and the availability of existing software using the approach which can be easily adapted for use with MOPoL (it is also the default MOBO acquisition function in the Python \textit{BoTorch} and \textit{Ax} packages \citep{facebook_ax_2022}).

qNEHVI is made up of three parts, expected hypervolume improvement (the EVHI), accounting for noisy data (the N) and the ability to produce several points in parallel (the q). On the first of these, qNEHVI seeks to maximise the hyper-volume of the Pareto frontier (i.e. the volume enclosed by the Pareto frontier for an arbitrary number of dimensions) by sampling hyperparameter values from its posterior distribution that it expects to yield the greatest improvement in hypervolume \citep{daulton_parallel_2021}. Essentially this is a generalisation of the single objective BO method of sampling based on expected improvement where settings which we estimate to have a high point estimate and a high variance are prioritised as this is where we would expect to find the most improvement. When it comes to accounting for noise, the acquisition function can integrate over a Gaussian noise term without wasting iterations trying to fit the noise as if it were signal. Finally, it can suggest several candidate hyperparameter settings it expects to jointly maximise the hypervolume improvement allowing for the assessment of several models in parallel.

The first step in the process is estimating doubly robust scores across all treatments and outcomes. In keeping with the optimal policy tree literature, this paper uses causal forest models to generate doubly-robust estimates. A multi-arm causal forest can be used for comparing multiple treatments rather than just treatment and control and it can also estimate the effects of these treatments on multiple outcomes \citep{athey_generalized_2019}.

The default surrogate function in the \textit{BoTorch} and the norm in BO more generally is Gaussian Process (GP) regression. GP incorporates the results of the evaluation function to calculate a posterior distribution updating the Pareto frontier and the uncertainty estimates around the frontier. This is by far the dominant approach for MOBO \citep{morales-hernandez_survey_2022}.

Now MOPoL begins the MOBO process finding the Pareto-frontier of non-dominated greedy trees or hybrid trees by estimating models on a hyperparameter or set of hyperparameters $\lambda \in \Lambda$ where $\Lambda$ specifies the characteristics defining the range of acceptable models in $\Pi$ (within the constraints placed on $\Pi$ in the previous section). These hyperparameters are acquired from qNEHVI’s acquisition function, starting from a random instantiation that samples that draws $2(N_y + 1)$ samples from a Sobol sequence where $N_y$ is the number of dimensions of the Pareto frontier \cite{daulton_parallel_2021}. These hyperparameters need not only be weightings for the outcome variables; any hyperparameter could be evaluated this way. Some particularly useful outcomes aside from expected value of various outcome variables might be the weighting of expected value for subgroups of interest, different settings for fairness adjustments (e.g. fairness regularisers) or different maximum depths (limiting the decision tree complexity). For non-continuous hyperparameters though, another distance metric e.g. Gower distance is necessary \citep{manson_mvmoo_2021}. Once the acquisition function has a candidate $\lambda$ it fits a surrogate model with the depth limited to the depth of the optimal tree candidates and expected value can be estimated from this set of rules for each outcome. Then standard errors are generated with bootstrapping. These estimates are added to the Pareto frontier if they are non-dominated. Finally, these quantities get plugged into the acquisition function to generate a new set of hyperparameters. This repeats until the process uses up its computation budget or until some other stopping criterion is reached.

\begin{figure}[!ht]
\centering
  \includegraphics[scale=0.7]{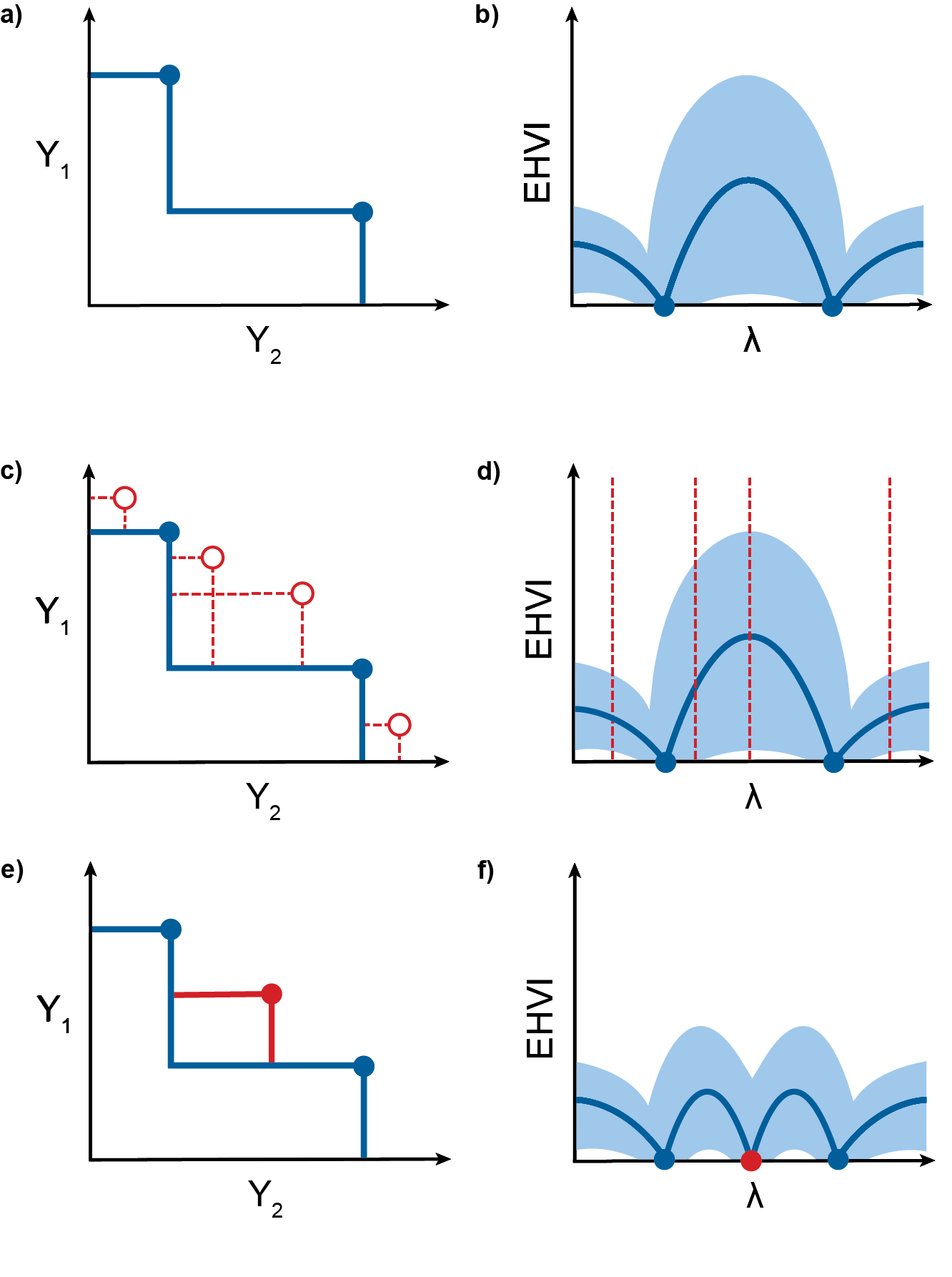}
  \caption{Stylised visualisation of the process of Expected Hypervolume Improvement MOBO}
  \label{fig:fig-stylised}
\end{figure}

Figure \ref{fig:fig-stylised} presents a stylised visualisation of what an Expected Hypervolume Improvement MOBO process does to fit a Pareto curve. The first row shows a MOBO process after having fit two points, Figure \ref{fig:fig-stylised}a is the Pareto frontier that is being formed with the boxed in area being the area under the Pareto frontier already. Figure \ref{fig:fig-stylised}b is the expected improvement for each value of a single hyperparameter of interest $\lambda$ with the point estimate as the dark blue line and the light blue as the uncertainty around the estimates. The second row shows some of the candidate values of $\lambda$ that the acquisition function considers. Figure \ref{fig:fig-stylised}c shows the results on the two outcomes that the function expects while Figure \ref{fig:fig-stylised}d shows the EHVI from each of these candidates. Note that area of the boxes in Figure \ref{fig:fig-stylised}c show the point estimate for the EHVI, the solid blue line in Figure \ref{fig:fig-stylised}d. The acquisition function selects the point corresponding to the maximum of the point estimate plus the uncertainty around it, it then evaluates this hyperparameter value and we see the result in the third row. The solid red point in Figure \ref{fig:fig-stylised}e has now been evaluated and added to the Pareto frontier and the solid red box has been added to the area. Note that the point didn't evaluate to precisely the expected outcome. Now the process will repeat. This is essentially the MOPoL process with the difference that qNEHVI integrates out noise in the Pareto frontier and can generate multiple candidate points that jointly maximise EHVI.

\begin{algorithm}
\caption{MOPoL algorithm for finding the Pareto frontier of surrogate models}\label{alg:cap}
\begin{algorithmic}
\Require Outcomes $Y$, covariates$X$, a set of possible hyperparameters $\Lambda$, an $\operatorname{AcquisitionFunction}$, a computational $budget$
\Ensure A Pareto set of models and hyperparameters $P$
\newline
\State $\hat{\Gamma} \gets$ estimate doubly robust scores for $Y$ with $X$
\newline
\State $i \gets 0$ 
\State Initialise empty Pareto set $P$
\State $\lambda_{init} \gets$ random sample $2(N_y + 1)$ points from a Sobol sequence of $\Lambda$
\State $\lambda_0 \gets \lambda_{init, \space 0}$ 
\State $start.time\gets$ current time
\While{$(start.time -$ current time$) < budget$}
\State $\pi_i \gets FitTree(\lambda, X,\hat{\Gamma})$ \hfill\Comment{Fit a greedy, hybrid or optimal policy tree}
\If{there is no model $\pi_j$ in $P$ for which $V_y(\pi_{j}) > V_y(\pi_{i})$ for all $y \in Y$}\hfill\Comment{update Pareto set}
\State Append $(\lambda_i, \pi_i)$ to $P$
\If{there is no model $\pi_j$ in the Pareto set for which $V_y(\pi_{j}) > V_y(\pi_{i})$ for all $y \in Y$}
\State Remove $(\lambda_j, \pi_j)$ from $P$
\EndIf
\EndIf
\If{$i < \operatorname{length} (\lambda_{init})$} \hfill\Comment{Check if initialisation points are exhausted}
\State $\lambda_{i+1} \gets \lambda_{init, \space i+1}$ \hfill\Comment{Iterate through initialisation points}
\Else
\State Initialise empty set of bootstrap value estimates $B$ of desired length
\For{$b$ in $1:\operatorname{length}(B)$ } \hfill\Comment{Bootstrap SE for acquisition function}
\State $X_{b} \gets $ boostrap sample $ X$
\State $\hat{\Gamma}_b \gets$ boostrap sample $\hat{\Gamma}$ 
\State $\pi_{b} \gets \operatorname{FitTree}(\lambda_i, X_b, \hat{\Gamma}_b)$
\State $v_b \gets V(\pi_b,\hat{\Gamma}_b)$
\EndFor
\State $SE_{V(\pi_i)} \gets \sqrt{\frac{\operatorname{Var}[v]}{\operatorname{length}(B)-1}}$
\State $\lambda_{i+1} \gets \operatorname{AcquisitionFunction}(V(\pi_i), SE_{V(\pi_i)})$ \hfill\Comment{Use acquisition function to get new point(s)}
\EndIf
\State $i \gets i + 1$
\EndWhile
\end{algorithmic}
\end{algorithm}

\FloatBarrier

From here, decision-makers can consider the Pareto frontier and then fit one or more optimal models. These optimal models can be compared directly on expected outcome measures or even on the nature of the decision rules to make a final choice.  For example, there will need to be a judgement on whether rules are fair and sufficiently intuitive. However, the Pareto frontier will have made choosing easier by mapping out the trade-offs between objectives.

\subsection{Causal identification}
While policy learning does not directly involve learning treatment effects, it does still rely on good causal identification in order to generate the doubly robust scores (or their equivalent in other policy learning approaches) necessary to fit a model \citep{athey_policy_2020}. For this reason the principles of good causal identification also apply to policy learning. The ideal scenario is to learn policies from large experimental data with high external validity. In this case, doubly robust inference adds little value and there is in theory no need to use doubly robust scores in learning at all, although using scores also should not be detrimental to the process. In observational contexts, we can still conduct policy learning but the usual caveats about modelling out selection effects in data apply. Essentially for the purposes of the \textit{grf}-based estimation we use in this paper, the $e(\cdot)$ and $m(\cdot)$ models that are fit have the same requirements as nuisance models that would be used in fitting treatment effect models \citep{nie_quasi-oracle_2020,chernozhukov_doubledebiased_2018}. This is a very important point because if confounders are not controlled for correctly, this will bias the policy tree in ways that cannot be diagnosed just looking at that final model alone. In economics it is typical to use approaches like difference-in-differences, instrumental variables or other approaches that attempt to design analysis that emulates a randomised controlled trial for better inference. We do not explore these approaches here because they are mostly concerned with generating unbiased local effects. This is useful for an evaluation, but less useful when trying to learn essentially heterogeneous treatment effects on which to base a set of rules that will apply to everyone.

\subsection{Obtaining standard errors}
Standard errors are useful in this case because they help the acquisition function to find areas with higher uncertainty and therefore greater EHVI. It is also useful to be able to see if there are regions on the Pareto frontier with large standard errors because these might be regions where we would be more sceptical about the ability of other samples to achieve similar values, i.e. that these points may be more likely to be dominated in other samples. Unfortunately, in the case of policy learning, there is no natural way to obtain standard errors that is less computationally intensive than non-parametric bootstrapping the expected value of a given tree for all outcomes. In practice this leads to a slow iteration time, particularly with the hybrid tree because so many models need to be fit.

In the application from this paper, we use a relatively small number of replicates by the standards of explanatory modelling. However, it may be possible to further cut down the number of replicates to speed up evaluation. The reason for this is that we are not using standard errors here for hypothesis testing, but rather for the purposes of quantifying the uncertainty around evaluation of a point. When it comes to the two different uses of the standard errors --- in defining uncertainty for human interpretation and in finding candidate hyperparameter values, the the benefit of prioritising smaller standard errors is not as clear as it is in causal inference. The human decision is much more of a heuristic judgement, likely based on relative sizes of standard errors; there is unlikely to be any hypothesis testing for example. When it comes to the acquisition function, the usefulness of standard errors is purely in acquiring points for estimating a Pareto frontier. The value of standard errors here is purely practical, and we have a clear utility function --- the hypervolume contained within the Pareto frontier to assess the trade-off between speed and goodness of fit. In addition speeding up the evaluation of each trial allows for more points to be assessed in a set computational budget, potentially improving hypervolume anyway.

Importantly, wide standard errors will not affect the point estimate of value, but they will cause the acquisition function to favour more uncertain points than it would under narrower errors. The reason for this is that because a MOBO algorithm only values upside uncertainty when picking points to evaluate, expanding standard errors by a certain proportion for all points would be expected to increase the extent to which qNEHVI samples more uncertain points. Given that it will still be making the same point estimates, it seems possible that for some questions it is well worth risking wider standard errors to allow for the evaluation of more points. Complicating this trade-off though is the fact that we would expect the runtime to evaluate a point to be roughly constant, whereas the acquisition function slows over time as more iterations are incorporated into the function. While there may be value to tuning the number of bootstrap replicates as a hyperparameter, it is probably simply better to make a judgement about the computational budget and the point at which the acquisition function begins to hit diminishing returns and then pick a number of replicates that reconciles these two factors.

\subsection{Parallelising MOPoL}
The last point worth discussing is parallelising MOPoL. As qNEHVI allows for the acquisition of an arbitrary number of hyperparameter settings at once. While parallel computing may speed up the fitting of a model, decision trees certainly do not benefit as much from being fit on the GPU as for example neural networks and it may even hurt fitting time in some cases. In theory it may be possible to achieve some speed-up by multiprocessing on the CPU through multi-processing (or by multi-threading if MOPoL were implemented in a language without a global interpreter lock). In the implementation used in this paper the algorithm draws a single point at a time and evaluates that point over a single thread. This was more than fast enough in the example given here (with one hyperparameter controlling two objectives), however, for fitting higher-dimensional frontiers, there is nothing stopping parallelisation in running multiple evaluation functions at once. This will likely speed up processing where evaluation models take a long time but as the complexity of the posterior increases and the acquisition time starts to be the main driver of slower iteration times, the benefits of parallelising will diminish.

\subsection{Deriving final policy rules}

Once a judgement has been made based on the Pareto frontier as to what the best trade-off between objectives is, a final set of rules needs to be created. There are numerous possible approaches. We would suggest fitting an optimal or hybrid tree, or a number of them to see the kind of rule-sets that are generated for a given weighting. A human decision maker should check over these and either approve one as the set of rules that will be used, or use them as the basis for handpicked rules. It may be useful to handpick rules for reasons of fairness or to make rules more easily understandable, for example, one might decide to make an age threshold 65 years old rather than 66.183 because the former is a commonly used cut-off that is easy to remember. The value generated by handpicked rules can then be calculated and compared with the Pareto frontier to understand how much potential value is being sacrificed so the rules can be handpicked.

Fitting the Pareto frontier did not use sample splitting to calculate out-of-sample error (because of the use of standard error). However, when finding an actual model in this final stage, it is important that standard machine learning practice of using a test set to estimate model performance should be followed. The topic of how to go about fitting this model is a standard policy learning problem (albeit with hyperparameter settings affecting the objective) and so will not be explored in great depth here. There should be no need to run MOPoL on a seperate validation set as is usually done for hyperparameter tuning in predictive machine learning because the choice of these weighting hyperparameters are not being used to improve performance on an objective function but rather to decide what objective function will be used. 

\FloatBarrier
\section{Application – Setting price subsidies for anti-malarials in Kenya} \label{sec:sec5application}

This application uses MOPoL to find treatment rules for allocating subsidies for anti-malarials in Kenya. This takes \citet{cohen_suppressing_2021} as its starting point and uses the replication data from that paper to try to design a set of rules for setting subsidies on a class of anti-malarial called artemisinin-based combination therapy (ACT). Malaria is a serious problem in Kenya. It causes deaths and suffering that are ultimately avoidable with good prevention and treatment. While government-run free clinics provide highly effective ACTs, free clinics can be inconvenient and suffer from shortages of medication. Many Kenyans instead get anti-malarial treatments from pharmacies. These are often closer, have longer opening hours and may not suffer from the same shortages. However, their products are not free and they sell a range of treatments which often means people buy cheaper, less effective medications.

Cohen et al. ran an RCT in four rural Kenyan areas in 2009. The goal of this program was to improve access to ACT treatment for those who had malaria. However, there was also concern about a high subsidy encouraging overuse of the drug. People may take the drug when they do not need to, for example prophylactically or to treat diseases that are not in fact malaria. Taking ACTs when they are not required can cause unneccessary shortages and accelerates the development of drug resistant malaria parasites \citep{cohen_price_2015}. Obviously, the primary concern here should be ensuring access to treatments for people whose lives may be at risk, but if as Cohen et al. suggest, there are easy gains to be made from imposing a higher price on some individuals (though one that still amounts to a heavy subsidy compared to the price of ACTs before the intervention) to prevent overuse, this is worth considering.

The data covered 2609 households and used 37 covariates for fitting doubly robust scores (listed in Appendix A). We excluded the no subsidy group due to the small sample size of just 173 households. While this may be an appropriately powered sample for the original analysis, it is simply too small to yield meaningful doubly robust scores for the purposes of exploring treatment heterogeneity via policy learning. Many propensity scores were close to 0 or 1 making them poorly suited for use in policy learning. In addition, the trade-off as framed by the original paper is one between differing high levels of subsidy, not subsidy or no subsidy, we could say the inclusion of no subsidy in the learner is not really policy relevant.

As with the original paper, we only examine the first pharmacist visit for anti-malarials for each household. This is because that is the only visit for which we can guarantee the household still had ACT vouchers. We can construct outcomes for these individuals based on whether they took ACTs when they had malaria (coded to 0 and 1) and whether they took ACTs when they did not have malaria (coded to 0 and -1 to make the value calculation intuitive). Unusually, we do not actually have ground truth for malaria diagnosis for most individuals who took an ACT during the period. Instead we have a probability score based on a predictive model developed by Cohen et al.. There are valid critiques of this model, but for the purposes of our application, we will take these estimates as accurate. We consider cases with $p < 0.3$ as being cases where the individual did not have malaria and cases where $p > 0.7$ as being cases where the individual did have malaria. While a more robust approach is probably to rescale these probabilities between $[-1,1]$ and use the probabilities to weight outcome values, we instead chose to binarise the outcome in order to demonstrate the use of this approach in the case where outcomes are binary which is much more common in applied work than a probability outcome.

We recode outcomes as follows where $Y$ is a recoded outcome and $A$ is a variable telling measuring whether the household bought ACTs from a study chemist: $$Y_{0,i}=A_i \cdot \#[Pr_i(malaria) > 0.7]$$ $$Y_{1,i}=-A_i \cdot \#[Pr_i(malaria) < 0.3].$$

\FloatBarrier

\subsection{Modelling doubly robust scores (and exploratory causal analysis)}

The first stage used a 20,000 tree multi-arm causal forest with two outcomes to estimate doubly robust scores for each of 2609 households. This means the nuisance functions were \textit{grf} regression forests with 20,000 trees and automatic hyperparameter tuning. The outcomes were standardised in-line with the recommendations made in the \textit{grf} documentation. Note that if the models to fit doubly robust scores fit a model and then estimate scores out-of-sample (like in \textit{grf}), it is not a problem that scores and policies are both learnt on the same sample of data or that we are not in some way cross-fitting the policy learner.

For the sake of getting intuition on the policy learner models, we present the average treatment effects calculated from the doubly robust scores and also variable importance from the causal forest (a depth-weighted count of the uses of each variable in tree splits over the ensemble).

\FloatBarrier

\begin{table}[ht]
\centering
\begin{tabular}{rllrr}
  \hline
 Outcome & Comparison & Estimate & SE \\ 
  \hline
    Correct use & 88\% - 92\% & 0.0807 & 0.0371 \\ 
    Correct use & 80\% - 92\% & -0.0191 & 0.0368 \\ 
    Incorrect use & 88\% - 92\% & 0.0236 & 0.0279 \\ 
    Incorrect use & 80\% - 92\% & 0.0350 & 0.0283 \\ 
   \hline
\end{tabular}
\caption{\label{ates}Average treatment effects by outcome and group (compared to Group 1)}
\end{table}

These results are not exactly what we might expect, in particular, the fact that lowering the subsidy to 88\% from 92\% supposedly results in a statistically significant increase in correct use of ACTs. This is not just a problem with the orthogonalisation models being inaccurate. The simple difference in means also shows a positive result, albeit one with smaller magnitude (a 0.037 difference). It is possible there is Giffen behaviour here, it is possible this is simply the result of sampling error. The rest of the estimates show the kind of results we would expect from Cohen et al.'s theoretical framework. However, the small magnitude of the effect on correct use for the 80\% subsidy suggests that there is a lot of elasticity in the effect of subsidies on correct use, i.e. there is room to impose a price on some cases that will decrease incorrect use and not affect (or even for some reason improve) correct use.

It is possible to get a sense of which variables correlate with variation in treatment effect --- and therefore which will be important in policy learning --- by examining the importance of different variables in constructing the causal forest. Variable importance can be understood as the percentage of variation in estimates that result from splitting on each variable across all the trees in the random forest. The optimal policy tree fit on these estimates will likely emulate the variable importance of the causal forest (as they are both learners trained on doubly robust scores), splitting more and higher in the tree on more important variables. Table \ref{var-imp} shows the age of the patient is by far the most important variable in predicting treatment effects, this makes sense given that age is a very important predictor of an illness being malaria in Cohen et al..

\begin{table}[ht]
\centering
\begin{tabular}{rlr}
  \hline
 Name& Label & Importance \\ 
  \hline
LOG\_patient\_age1 & admin data: age of patient, 1st log entry for the hh & 0.4979 \\ 
  B\_antimal\_cost & average cost across malaria episodes for which cost provided & 0.0547 \\ 
  dist\_clinic & distance to closest health center & 0.0539 \\ 
  head\_acres & acres of land & 0.0521 \\ 
  B\_dist\_km & distance (km) from household to study chemist & 0.0465 \\ 
  B\_adultteen & number of adult and teen members (age$>$=9)  in hh & 0.0440 \\ 
  head\_edu & years of education of household head & 0.0383 \\ 
  B\_head\_age & age of head & 0.0287 \\ 
  share\_undernet & baseline: share of hh members sleeping under net (computed) & 0.0237 \\ 
  B\_mal\_ct & baseline table C: total malaria episodes in hh & 0.0237 \\ 
   \hline
\end{tabular}
\caption{\label{var-imp}Variable importances for top 10 variables}
\end{table}
\FloatBarrier

\subsection{Estimating a Pareto frontier}

\begin{figure}[!ht]
  \includegraphics[]{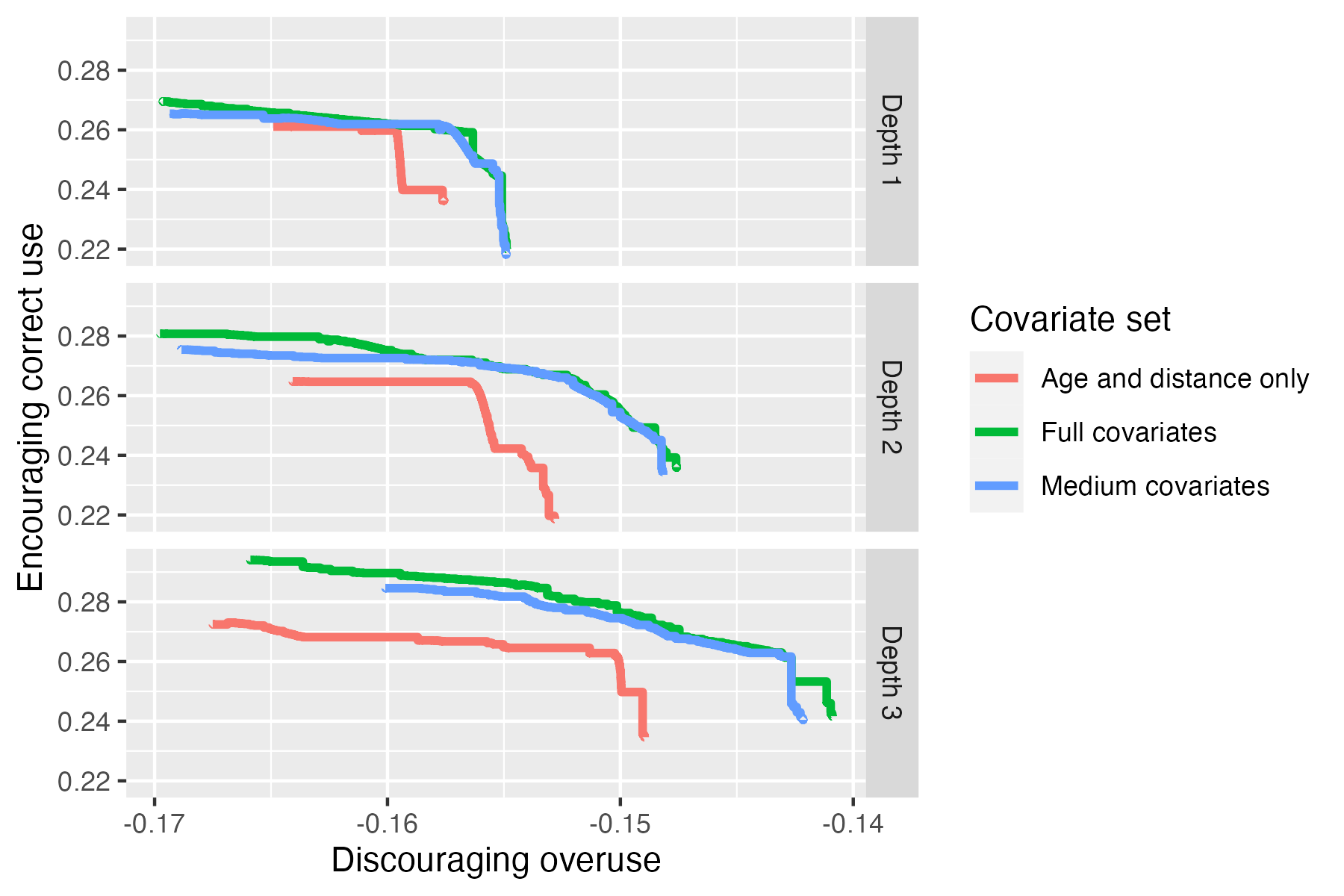}
  \caption{Greedy, hybrid and oracle Pareto frontier outcomes for training data}
  \label{fig:fig2}
\end{figure}

We estimate a Pareto frontier across values of the hyperparameter $\lambda \in [0,1]$ using MOPoL. Figure \ref{fig:fig2} shows the Pareto frontiers for trees of depths 1, 2 and 3.

In this case we see that the Pareto frontier shows relatively large marginal gains in preventing misuse can be made from a small increase in price for some groups up until an elbow point where further gains come at the expense of much larger loss in access. Deeper decision trees seem to yield better outcomes particularly on preventing misuse.

\begin{figure}
  \includegraphics[scale=0.7]{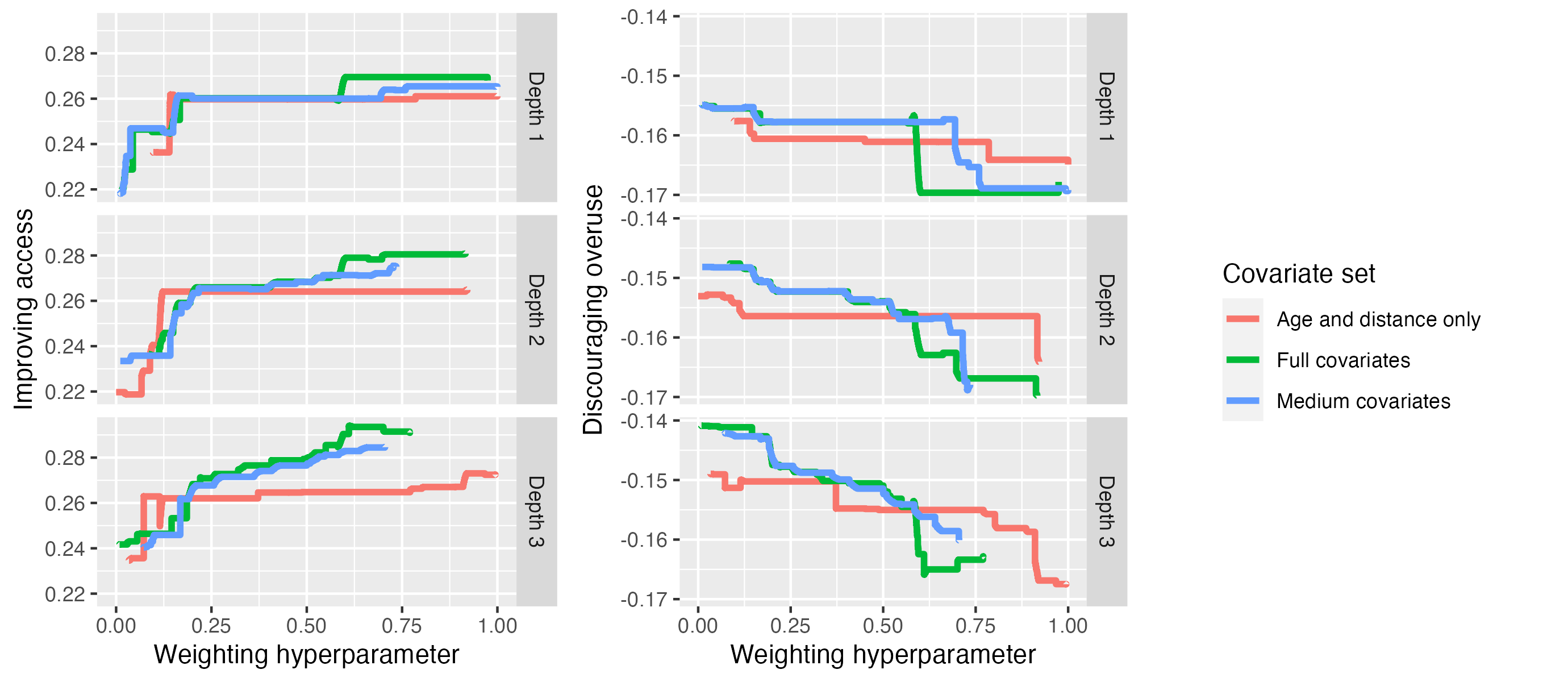}
  \caption{Change in expected outcome for encouraging correct use across weighting parameter values}
  \label{fig:fig4}
\end{figure}

It is also useful to look at expected outcomes across values of $\lambda$. Figure \ref{fig:fig4} lays this out across both outcomes and show a similar pattern to the one show in Figure \ref{fig:fig2}, relatively flat values for high weightings of the outcome, followed by a drop-off in the lower end of the hyperparameter values. The curve is particularly flat for the models fit just on age and distance suggesting that while there is little cost in access for reducing subsidies, there is also not much benefit for doing so. Note that some of the frontiers do not continue all the way to the axis. This shows the limits of the hyperparameter drawn assessed by the acquisition function. The function found it so unlikely that the outcomes for these areas would be higher than they were for the values that were already assessed that it did not judge these worth sampling.

These three figures can be used as a basis for weighting outcomes. Intuitively, if we value access over correct use, we would likely want to set the hyperparameter value somewhere in the region $0.5 \leq \lambda \leq 1.0$ where the increase in prices has not affected access but has reduced incorrect use somewhat.

\FloatBarrier

\subsection{Fitting a final optimal policy tree}

The final stage involves fitting the optimal policy tree based on a chosen hyperparameter. This analysis uses $\lambda= 0.5$ as a good trade-off between the objectives. We fit trees for each of the sets of covariates and for depths 2 and 3. This is partly to see the performance effects of the different depths and data size and partly to be able to look across these models and see which seem to be reasonable rules for discrimination and which seem unreasonable.

\begin{figure}[!ht]
\begin{tabular}{ccc}
 & Depth 2 & Depth 3 \\ \\
  All variables & 
  \includegraphics[width=40mm]{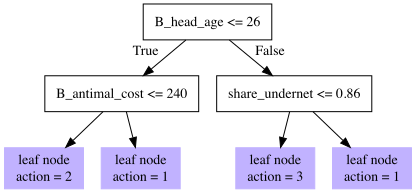} &   \includegraphics[width=80mm]{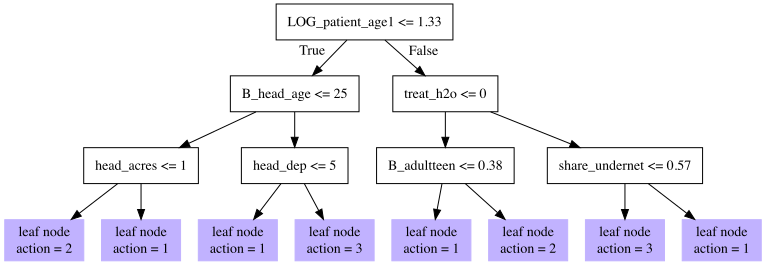} \\
  Selected variables & 
  \includegraphics[width=40mm]{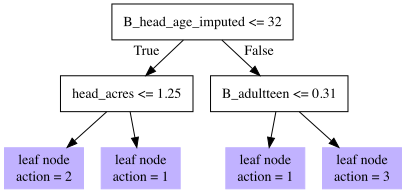} &
  \includegraphics[width=80mm]{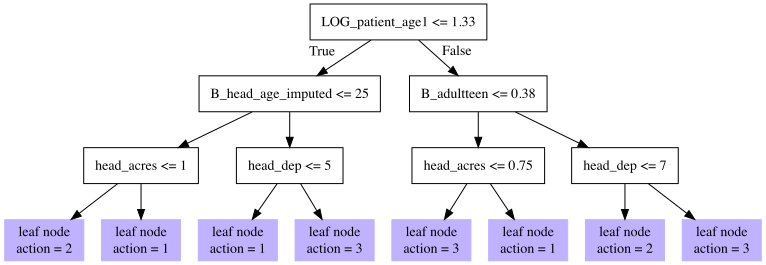} \\
  \shortstack{Age and distance \\ from clinic} &
  \includegraphics[width=40mm]{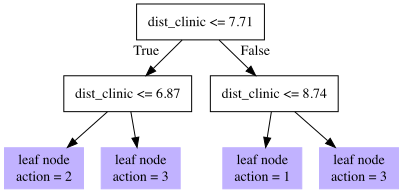} &
  \includegraphics[width=80mm]{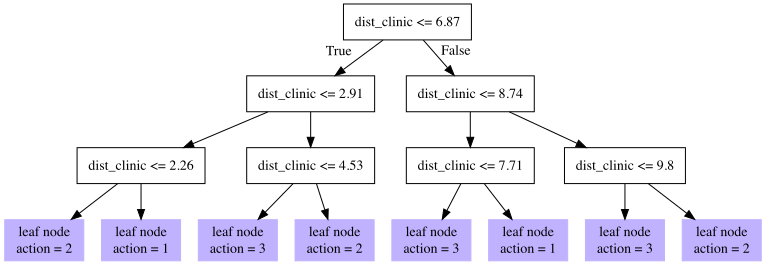} \\
\end{tabular}
\label{fig:final-trees}
\caption{Final policy trees with varying settings}
\end{figure}

Figure \ref{fig:final-trees} shows optimal trees fit on 1305 households with 1304 held out as a test sample. The value for each outcome for the test and train samples and the time taken to fit each model is shown in Figure \ref{fig:final-trees}. There are a few take-aways from this. While the full set of covariates yields marginally better value, the medium and small sets seem to be more the kinds of models that make policy sense. For example, the variables are not about knowledge or manipulable characteristics (like the number of sleepers under mosquito nets). The small set of covariates yields worse value but the rules are more intuitive and the covariates used seem to give the most just explanation for the rules. Children are most likely to benefit from the treatment and anyone who lives close to a clinic but goes to the pharmacy anyway may be using the medication in such a way that a clinic will not give them any. 

A particularly striking result is that the cost of an extra layer of depth in terms of performance is dramatic. While we would not want a tree much deeper than three splits for most applications because it undermines the comprehensibility of the rules, it seems like a tree any deeper would likely be completely infeasible anyway, at least for this dataset. A hybrid tree would likely be a better alternative for a deeper tree. In addition, the poor scaling of the optimal tree in terms of number of variables can be seen as well in the extreme differences in fit time for different sets of variables.

\begin{table}[t]
\centering
\begin{tabular}{rrrrrrrr}
  \hline
  Depth &
  Variables &
 \shortstack{Train value -\\ Improved access} 
 & \shortstack{Test value -\\ Improved access} & 
 \shortstack{Train value -\\ \shortstack{Discouraging\\overuse}} & 
 \shortstack{Test value -\\ \shortstack{Discouraging\\overuse}} &
 \shortstack{Fit time \\ (seconds)}\\ 
  \hline
2 & All variables & 0.259 & 0.251 & -0.128 & -0.162 & 1.197 \\ 
2 & Selected variables & 0.256 & 0.238 & -0.130 & -0.173 & 0.130 \\ 
2 & Age and distance from clinic & 0.251 & 0.252 & -0.139 & -0.155 & 0.021 \\ 
3 & All variables & 0.275 & 0.251 & -0.112 & -0.166 & 2953.669 \\ 
3 & Selected variables & 0.274 & 0.245 & -0.113 & -0.170 & 137.349 \\ 
3 & Age and distance from clinic & 0.266 & 0.262 & -0.126 & -0.154 & 19.572 \\ 
   \hline
\end{tabular}
\caption{Value results from final policy trees fit with various settings for $\lambda=0.5$.\\NB: Models were fit on a 2023 MacBook Air with a 3.49GHz, eight core M2 processor.}
\end{table}

As a caveat on these findings, it is worth being sceptical about these findings and their external validity. These results are in conflict with the theory presented by \citet{cohen_price_2015} for example in the effect from moving to a 92\% to an 88\% subsidy. If this were an actual policy to be put in place governing the access that very poor people would have to life-saving medical interventions, we would want to have much more confidence in the analysis than what this small field experiment can give us. Obviously, an actual study to design a policy should also look at the effect of lower subsidies on economic hardship as well as just demand for ACTs. While it is a good thing that we can use price to decrease misuse of the drug with only a small decrease in proper use, it is not just a problem if people who need care are priced out of receiving it, it is also a problem if people who need care are harmed financially by paying for treatment.
In addition, it would be worth understanding the mechanisms behind non-intuitive findings to get a better sense of whether this is actually causation or just some correlation that we would not expect to see when generalising out the policy rules. For example, an explanatory mixed-methods design would be useful here \citep{plano_clark_mixed_2022}. It is also possible that some of these counterintuitive findings come from the original study being essentially underpowered for policy learning such that the policy learnt is mostly noise even if it were correctly powered for inference about the average treatment effect in Cohen et al..

\FloatBarrier

\subsection{Exploring different configurations}

Finally, in order to get a sense of where improvements could be made in designing a MOPoL study for this problem, we test a variety of different specifications. There are three aspects of the design we are interested in for this section, firstly the number of bootstraps used to fit standard errors, secondly the benefits of increasing the number of iterations and the increasing marginal time cost of each iteration, and finally the use of a decision tree which is optimal or greedy. In this section, all trees are fit to depth two and by default use 100 bootstrap replicates for standard errors. We test these looking at fitting time on this dataset and also the area under the Pareto frontier. This area is not in itself particularly important, instead we care about the ability of the frontier to characterise the trade-off between objectives. However, area (and more generally hypervolume) is a good proxy for a frontier where coverage across different hyperparameters is good and where we are trying to approximate an optimal frontier which would by definition sit at or above the greedy frontier at all points.

While the findings from this section may not generalise across problems, these results will hopefully help to demonstrate how specifications can be tested to yield the best possible results within a computational budget. Importantly, this is a simple two objective case where a single hyperparameter bounded $\lambda \in [0,1]$ is being used to control outcomes. For problems with higher dimensionality and potentially a larger range for hyperparameters to take, we might expect more compute would be needed (particularly a higher number of iterations).

\begin{figure}[!ht]
  \includegraphics[]{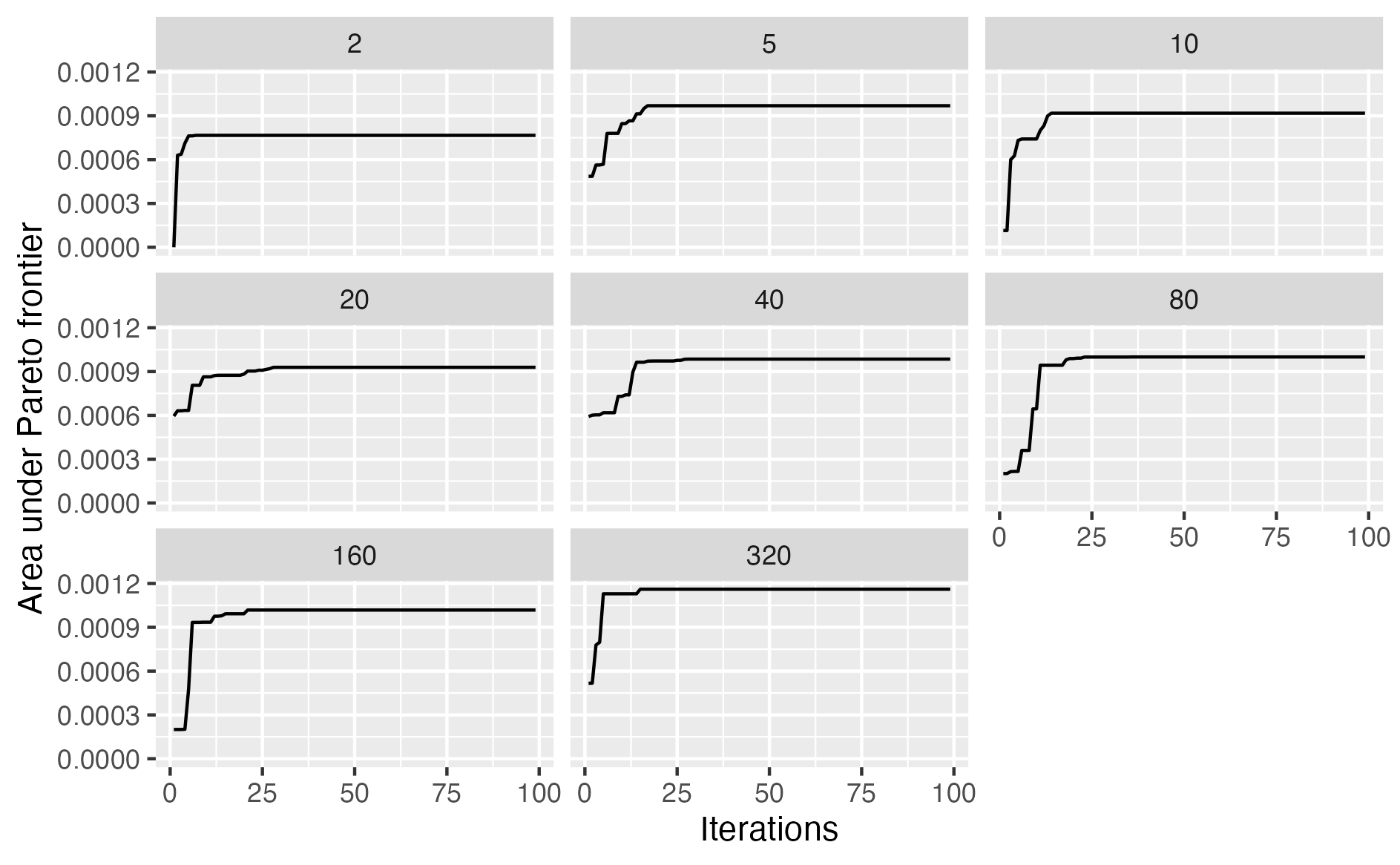}
\caption{The area within the Pareto frontier for different numbers of bootstrap replicates across 100 iterations of MOPoL with a greedy tree.}
\label{fig:bootstraps}
\end{figure}

Figure \ref{fig:bootstraps} explores the effect that the number of bootstrap replicates has on fitting of the Pareto frontier while Table \ref{tab:bs-it-time} shows the fitting time. A higher number of bootstrap replicates causes the volume to plateau higher and seemingly to reach that plateau sooner as well. The intuition on the lower Pareto frontier for lower numbers of replicates is that essentially high variance from some very unreliable standard errors concentrates evaluations in certain areas by chance given the acquisition function is drawn to areas with higher variance. Because of that, we do not evaluate some areas of the frontier which would help to make the frontier more convex, therefore enclose more area. However, larger numbers of bootstraps are more time intensive and so there is a trade-off to be made here to be considered alongside other possible efficiency gains in this section. In this case the amount of time needed though does not scale fast enough to be impractical. In fact, because the acquisition function slows down iterations, we see using two replicates per iteration ends up being speeding up fitting by a factor of less than $10 \times$ over a frontier fit with 320 replicates.

\begin{table}[]
\centering
\begin{tabular}{ll}
\hline \\
Number of replicates & Run time (seconds)\\
\hline \\
2                    & 145.348\\
5                    & 158.027\\
10                   & 170.975\\
20                   & 203.097\\
40                   & 281.050\\
80                   & 427.991\\
160                  & 716.711\\
320                  & 1228.253\\
\hline
\end{tabular}
\caption{Time taken to fit each Pareto frontier for testing effect of bootstrap replicates}
\label{tab:bs-it-time}
\end{table}

\begin{figure}[!ht]
  \includegraphics[]{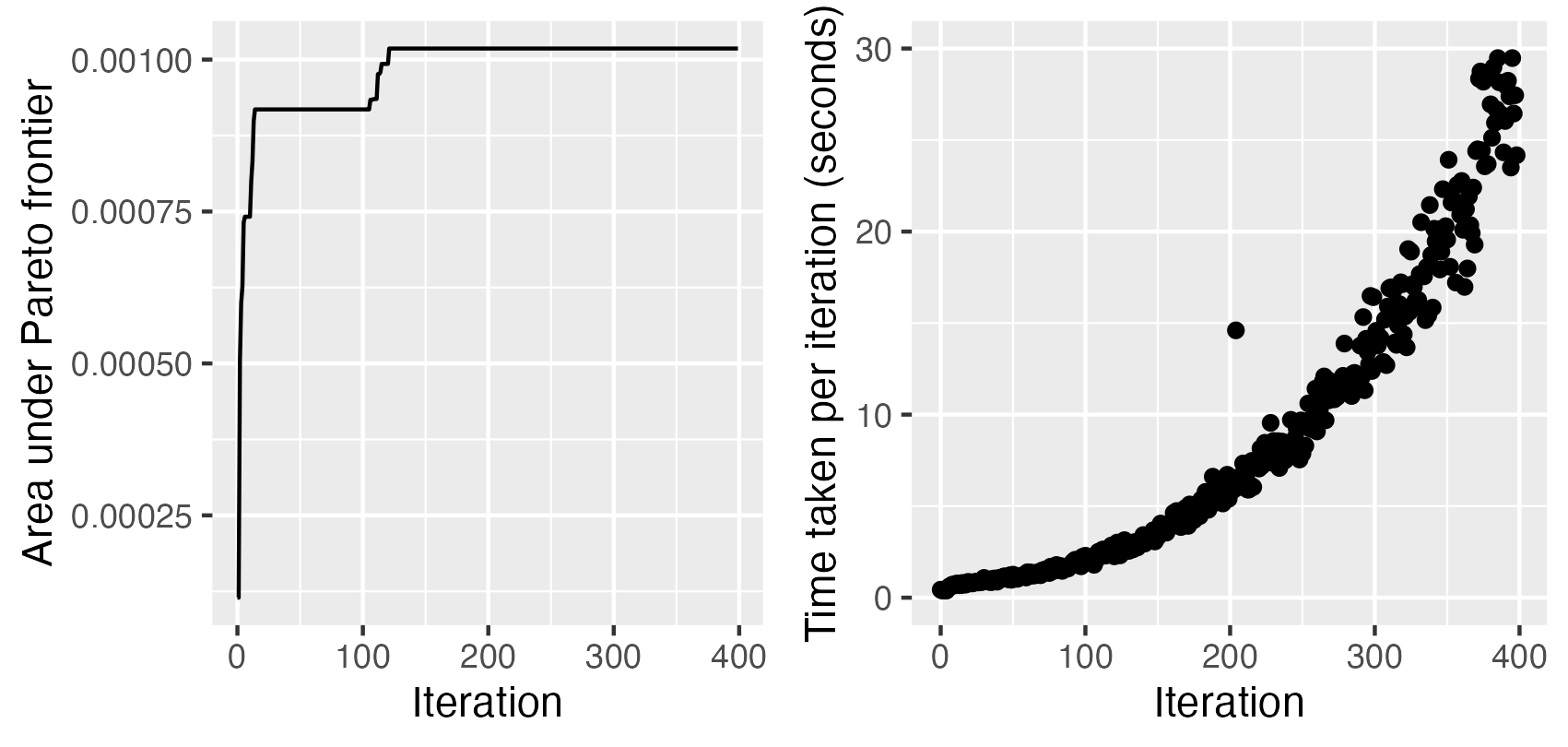}
  \caption{Area under the Pareto frontier with each iterations (upper) and time (in seconds) for each iteration (lower) for 400 iterations.}
  \label{fig:dim_iter}
\end{figure}

Figure \ref{fig:dim_iter} examines the effect of the number of iterations on area under the Pareto frontier and the time taken per iteration (here it is the acquisition function that begins to c. There are rapid gains in the first few iterations of the model and then a plateauing after that. In addition, the acquisition function becomes more and more time consuming with more iterations meaning that eventually we are using a lot of computation to find a point with only marginal (if any) gains.

\begin{figure}[!ht]
  \includegraphics[]{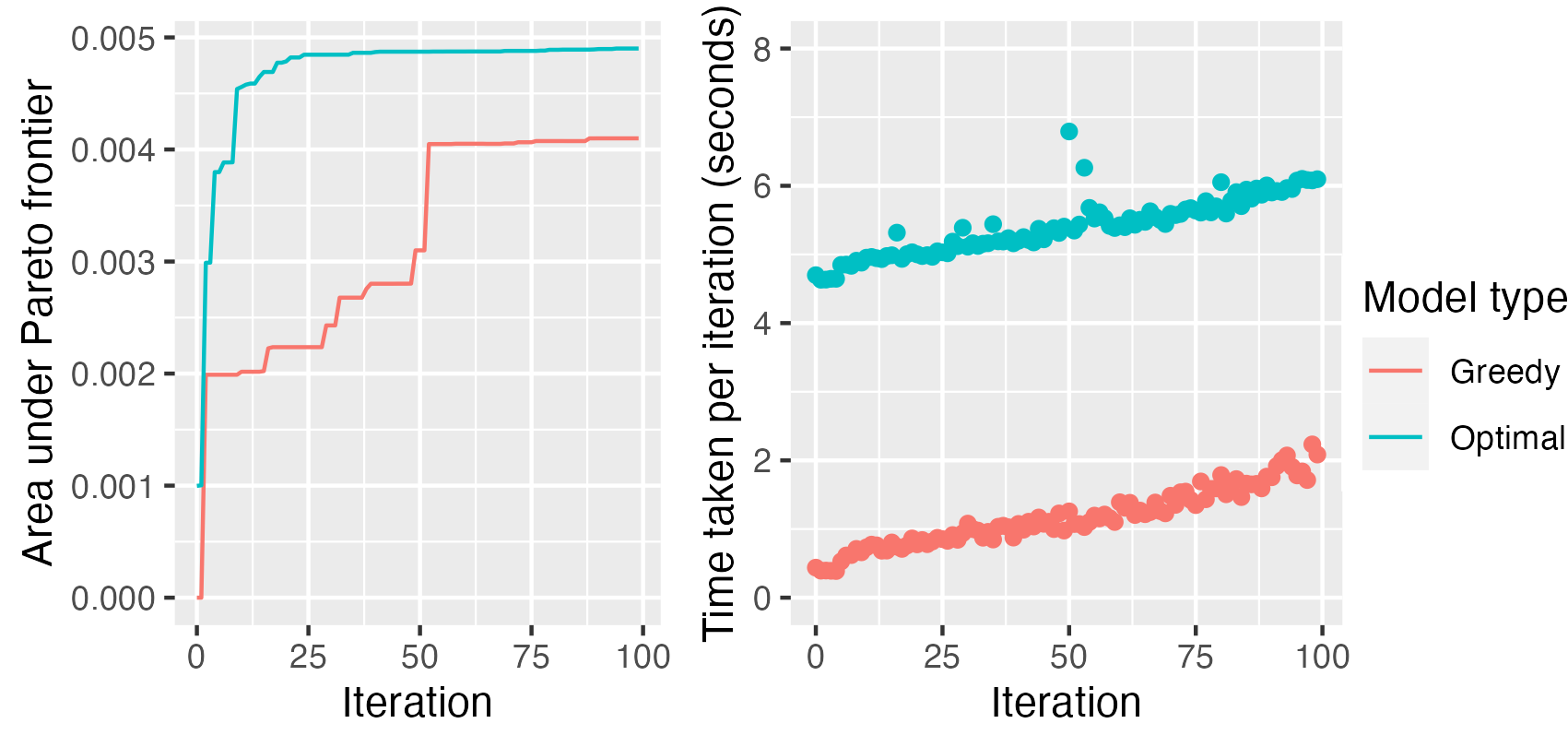}
  \caption{Area under the Pareto frontier with each iterations (upper) and time (in seconds) for each iteration (lower) by model type for 100 iterations.}
  \label{fig:opt_v_greedy}
\end{figure}

Figure \ref{fig:opt_v_greedy} shows the benefits of using a tree which optimised across two splits at a time rather than one (in this case fully optimal tree as the depth was only two). The optimal tree takes many times as long to fit as the greedy tree as can be seen in the gap in times at the first iteration. The time to fit a tree is constant across iterations while the acquisition function increases over time so this gap is simply the differences in the evaluation models. Because of this, even though the greedy tree is substantially faster initially, over time, the ratio between the two moves closer to 1 as the acquisition function begins to take up a greater portion of the time for each iteration.

\section{Discussion}

Using multi-objective Bayesian optimisation can help to design rules for allocating different treatments in situations where decision-makers are interested in multiple outcomes. It creates the flexibility to make decisions about the weightings of different objectives based on a fairly accurate estimation of how different optimal models will perform, without having to waste computing power on fitting a large number of optimal models.

While this paper has proposed a basic version of the MOPoL, there is still more to be done to develop the method and test it in other applications. Final model value and computational overhead could both be improved (or better balanced) by exploring a range of different approaches to MOBO. In particular, there are three design decisions that need to be better assessed as all have a computational cost. Greedy vs hybrid vs optimal models, the number of bootstrap replicates used and the sample size. As mentioned in Section \ref{choosing-eval-fn}, there is also a case to be made for ignoring optimal trees in larger sample sizes as fitting such a model is simply too time complex to be practical and so, we would likely achieve better results by fitting a less time complex model on a larger sample. In addition, more bootstrap replicates in standard error calculation with a cheaper model may be a better way to fit a Pareto frontier. Future work is needed to better understand these design trade-offs.

One useful development for the method could be extending it to the problem of optimisation under a budget constraint. While this paper focused only on unconstrained optimisation, there is no reason why applying a budget constraint to optimal trees and surrogate functions would not be possible \citep{verhaeghe_learning_2020}. In an application like this, a tree could give the best allocation of a scarce good and a soft budget constraint itself could even be one of the hyperparameters being tuned.

The suitability of using MOPoL in different areas of decision-making depends on the stakes of decisions, social license, transparency needs and quantifiability of outcomes \citep{mittelstadt_ethics_2016,ireni_saban_ethical_2022}. In cases where turning decisions over to the algorithm entirely would be inappropriate, a Pareto frontier still might be useful to check human-made rules against the set of models which dominate the human-made rules to see if there are sensible changes that could be made to achieve better expected value. Here, one would calculate a Pareto frontier, find the point on it which best matches the revealed weightings of outcomes by the human policymaker then calculate regret using the optimal tree as a reference.

\bibliographystyle{apacite}
\bibliography{references}  

\begin{thebibliography}{}

\bibitem [\protect \citeauthoryear {%
Amram%
, Dunn%
\BCBL {}\ \BBA {} Zhuo%
}{%
Amram%
\ \protect \BOthers {.}}{%
{\protect \APACyear {2020}}%
}]{%
amram_optimal_2020}
\APACinsertmetastar {%
amram_optimal_2020}%
\begin{APACrefauthors}%
Amram, M.%
, Dunn, J.%
\BCBL {}\ \BBA {} Zhuo, Y\BPBI D.%
\end{APACrefauthors}%
\unskip\
\newblock
\APACrefYearMonthDay{2020}{{\APACmonth{12}}}{}.
\newblock
\APACrefbtitle {Optimal {Policy} {Trees}.} {Optimal {Policy} {Trees}.}
\newblock
\APACaddressPublisher{}{arXiv}.
\newblock
\begin{APACrefURL} [{2022-08-18}]\url{http://arxiv.org/abs/2012.02279} \end{APACrefURL}
\newblock
\APACrefnote{arXiv:2012.02279 [cs]}
\PrintBackRefs{\CurrentBib}

\bibitem [\protect \citeauthoryear {%
Athey%
, Tibshirani%
\BCBL {}\ \BBA {} Wager%
}{%
Athey%
\ \protect \BOthers {.}}{%
{\protect \APACyear {2019}}%
}]{%
athey_generalized_2019}
\APACinsertmetastar {%
athey_generalized_2019}%
\begin{APACrefauthors}%
Athey, S.%
, Tibshirani, J.%
\BCBL {}\ \BBA {} Wager, S.%
\end{APACrefauthors}%
\unskip\
\newblock
\APACrefYearMonthDay{2019}{{\APACmonth{04}}}{}.
\newblock
{\BBOQ}\APACrefatitle {Generalized random forests} {Generalized random forests}.{\BBCQ}
\newblock
\APACjournalVolNumPages{The Annals of Statistics}{47}{2}{1148--1178}.
\newblock
\begin{APACrefURL} \url{https://doi.org/10.1214/18-AOS1709} \end{APACrefURL}
\newblock
\begin{APACrefDOI} \doi{10.1214/18-AOS1709} \end{APACrefDOI}
\PrintBackRefs{\CurrentBib}

\bibitem [\protect \citeauthoryear {%
Athey%
\ \BBA {} Wager%
}{%
Athey%
\ \BBA {} Wager%
}{%
{\protect \APACyear {2020}}%
}]{%
athey_policy_2020}
\APACinsertmetastar {%
athey_policy_2020}%
\begin{APACrefauthors}%
Athey, S.%
\BCBT {}\ \BBA {} Wager, S.%
\end{APACrefauthors}%
\unskip\
\newblock
\APACrefYearMonthDay{2020}{{\APACmonth{09}}}{}.
\newblock
\APACrefbtitle {Policy {Learning} with {Observational} {Data}.} {Policy {Learning} with {Observational} {Data}.}
\newblock
\APACaddressPublisher{}{arXiv}.
\newblock
\begin{APACrefURL} [{2022-08-18}]\url{http://arxiv.org/abs/1702.02896} \end{APACrefURL}
\newblock
\APACrefnote{arXiv:1702.02896 [cs, econ, math, stat]}
\PrintBackRefs{\CurrentBib}

\bibitem [\protect \citeauthoryear {%
Bertsimas%
\ \BBA {} Dunn%
}{%
Bertsimas%
\ \BBA {} Dunn%
}{%
{\protect \APACyear {2017}}%
}]{%
bertsimas_optimal_2017}
\APACinsertmetastar {%
bertsimas_optimal_2017}%
\begin{APACrefauthors}%
Bertsimas, D.%
\BCBT {}\ \BBA {} Dunn, J.%
\end{APACrefauthors}%
\unskip\
\newblock
\APACrefYearMonthDay{2017}{{\APACmonth{07}}}{}.
\newblock
{\BBOQ}\APACrefatitle {Optimal classification trees} {Optimal classification trees}.{\BBCQ}
\newblock
\APACjournalVolNumPages{Machine Learning}{106}{7}{1039--1082}.
\newblock
\begin{APACrefURL} \url{https://doi.org/10.1007/s10994-017-5633-9} \end{APACrefURL}
\newblock
\begin{APACrefDOI} \doi{10.1007/s10994-017-5633-9} \end{APACrefDOI}
\PrintBackRefs{\CurrentBib}

\bibitem [\protect \citeauthoryear {%
Bertsimas%
, Dunn%
\BCBL {}\ \BBA {} Mundru%
}{%
Bertsimas%
\ \protect \BOthers {.}}{%
{\protect \APACyear {2019}}%
}]{%
bertsimas_optimal_2019}
\APACinsertmetastar {%
bertsimas_optimal_2019}%
\begin{APACrefauthors}%
Bertsimas, D.%
, Dunn, J.%
\BCBL {}\ \BBA {} Mundru, N.%
\end{APACrefauthors}%
\unskip\
\newblock
\APACrefYearMonthDay{2019}{{\APACmonth{04}}}{}.
\newblock
{\BBOQ}\APACrefatitle {Optimal {Prescriptive} {Trees}} {Optimal {Prescriptive} {Trees}}.{\BBCQ}
\newblock
\APACjournalVolNumPages{INFORMS Journal on Optimization}{1}{2}{164--183}.
\newblock
\begin{APACrefURL} [{2022-11-21}]\url{http://pubsonline.informs.org/doi/10.1287/ijoo.2018.0005} \end{APACrefURL}
\newblock
\begin{APACrefDOI} \doi{10.1287/ijoo.2018.0005} \end{APACrefDOI}
\PrintBackRefs{\CurrentBib}

\bibitem [\protect \citeauthoryear {%
Binder%
, Moosbauer%
, Thomas%
\BCBL {}\ \BBA {} Bischl%
}{%
Binder%
\ \protect \BOthers {.}}{%
{\protect \APACyear {2020}}%
}]{%
binder_multi-objective_2020}
\APACinsertmetastar {%
binder_multi-objective_2020}%
\begin{APACrefauthors}%
Binder, M.%
, Moosbauer, J.%
, Thomas, J.%
\BCBL {}\ \BBA {} Bischl, B.%
\end{APACrefauthors}%
\unskip\
\newblock
\APACrefYearMonthDay{2020}{{\APACmonth{06}}}{}.
\newblock
{\BBOQ}\APACrefatitle {Multi-objective hyperparameter tuning and feature selection using filter ensembles} {Multi-objective hyperparameter tuning and feature selection using filter ensembles}.{\BBCQ}
\newblock
\BIn{} \APACrefbtitle {Proceedings of the 2020 {Genetic} and {Evolutionary} {Computation} {Conference}} {Proceedings of the 2020 {Genetic} and {Evolutionary} {Computation} {Conference}}\ (\BPGS\ 471--479).
\newblock
\APACaddressPublisher{Cancún Mexico}{ACM}.
\newblock
\begin{APACrefURL} [{2022-11-22}]\url{https://dl.acm.org/doi/10.1145/3377930.3389815} \end{APACrefURL}
\newblock
\begin{APACrefDOI} \doi{10.1145/3377930.3389815} \end{APACrefDOI}
\PrintBackRefs{\CurrentBib}

\bibitem [\protect \citeauthoryear {%
Breiman%
}{%
Breiman%
}{%
{\protect \APACyear {2001}}%
}]{%
breiman_statistical_2001}
\APACinsertmetastar {%
breiman_statistical_2001}%
\begin{APACrefauthors}%
Breiman, L.%
\end{APACrefauthors}%
\unskip\
\newblock
\APACrefYearMonthDay{2001}{{\APACmonth{08}}}{}.
\newblock
{\BBOQ}\APACrefatitle {Statistical {Modeling}: {The} {Two} {Cultures} (with comments and a rejoinder by the author)} {Statistical {Modeling}: {The} {Two} {Cultures} (with comments and a rejoinder by the author)}.{\BBCQ}
\newblock
\APACjournalVolNumPages{Statistical Science}{16}{3}{199--231}.
\newblock
\begin{APACrefURL} \url{https://doi.org/10.1214/ss/1009213726} \end{APACrefURL}
\newblock
\begin{APACrefDOI} \doi{10.1214/ss/1009213726} \end{APACrefDOI}
\PrintBackRefs{\CurrentBib}

\bibitem [\protect \citeauthoryear {%
Breiman%
, Friedman%
, Olshen%
\BCBL {}\ \BBA {} Stone%
}{%
Breiman%
\ \protect \BOthers {.}}{%
{\protect \APACyear {2017}}%
}]{%
breiman_classification_2017}
\APACinsertmetastar {%
breiman_classification_2017}%
\begin{APACrefauthors}%
Breiman, L.%
, Friedman, J\BPBI H.%
, Olshen, R\BPBI A.%
\BCBL {}\ \BBA {} Stone, C\BPBI J.%
\end{APACrefauthors}%
\unskip\
\newblock
\APACrefYear{2017}.
\newblock
\APACrefbtitle {Classification {And} {Regression} {Trees}} {Classification {And} {Regression} {Trees}}\ (\PrintOrdinal{1}\ \BEd).
\newblock
\APACaddressPublisher{}{Routledge}.
\newblock
\begin{APACrefURL} [{2022-11-21}]\url{https://www.taylorfrancis.com/books/9781351460491} \end{APACrefURL}
\newblock
\begin{APACrefDOI} \doi{10.1201/9781315139470} \end{APACrefDOI}
\PrintBackRefs{\CurrentBib}

\bibitem [\protect \citeauthoryear {%
Carvalho%
, Feller%
, Murray%
, Woody%
\BCBL {}\ \BBA {} Yeager%
}{%
Carvalho%
\ \protect \BOthers {.}}{%
{\protect \APACyear {2019}}%
}]{%
carvalho_assessing_2019}
\APACinsertmetastar {%
carvalho_assessing_2019}%
\begin{APACrefauthors}%
Carvalho, C.%
, Feller, A.%
, Murray, J.%
, Woody, S.%
\BCBL {}\ \BBA {} Yeager, D.%
\end{APACrefauthors}%
\unskip\
\newblock
\APACrefYearMonthDay{2019}{{\APACmonth{09}}}{}.
\newblock
{\BBOQ}\APACrefatitle {Assessing {Treatment} {Effect} {Variation} in {Observational} {Studies}: {Results} from a {Data} {Challenge}} {Assessing {Treatment} {Effect} {Variation} in {Observational} {Studies}: {Results} from a {Data} {Challenge}}.{\BBCQ}
\newblock
\APACjournalVolNumPages{arXiv:1907.07592 [stat]}{}{}{}.
\newblock
\begin{APACrefURL} [{2021-08-20}]\url{http://arxiv.org/abs/1907.07592} \end{APACrefURL}
\newblock
\APACrefnote{arXiv: 1907.07592}
\PrintBackRefs{\CurrentBib}

\bibitem [\protect \citeauthoryear {%
Chernozhukov%
\ \protect \BOthers {.}}{%
Chernozhukov%
\ \protect \BOthers {.}}{%
{\protect \APACyear {2018}}%
}]{%
chernozhukov_doubledebiased_2018}
\APACinsertmetastar {%
chernozhukov_doubledebiased_2018}%
\begin{APACrefauthors}%
Chernozhukov, V.%
, Chetverikov, D.%
, Demirer, M.%
, Duflo, E.%
, Hansen, C.%
, Newey, W.%
\BCBL {}\ \BBA {} Robins, J.%
\end{APACrefauthors}%
\unskip\
\newblock
\APACrefYearMonthDay{2018}{{\APACmonth{02}}}{}.
\newblock
{\BBOQ}\APACrefatitle {Double/debiased machine learning for treatment and structural parameters} {Double/debiased machine learning for treatment and structural parameters}.{\BBCQ}
\newblock
\APACjournalVolNumPages{The Econometrics Journal}{21}{1}{C1--C68}.
\newblock
\begin{APACrefURL} [{2021-05-13}]\url{https://doi.org/10.1111/ectj.12097} \end{APACrefURL}
\newblock
\begin{APACrefDOI} \doi{10.1111/ectj.12097} \end{APACrefDOI}
\PrintBackRefs{\CurrentBib}

\bibitem [\protect \citeauthoryear {%
J.~Cohen%
, Dupas%
\BCBL {}\ \BBA {} Schaner%
}{%
J.~Cohen%
\ \protect \BOthers {.}}{%
{\protect \APACyear {2015}}%
}]{%
cohen_price_2015}
\APACinsertmetastar {%
cohen_price_2015}%
\begin{APACrefauthors}%
Cohen, J.%
, Dupas, P.%
\BCBL {}\ \BBA {} Schaner, S.%
\end{APACrefauthors}%
\unskip\
\newblock
\APACrefYearMonthDay{2015}{{\APACmonth{02}}}{}.
\newblock
{\BBOQ}\APACrefatitle {Price {Subsidies}, {Diagnostic} {Tests}, and {Targeting} of {Malaria} {Treatment}: {Evidence} from a {Randomized} {Controlled} {Trial}} {Price {Subsidies}, {Diagnostic} {Tests}, and {Targeting} of {Malaria} {Treatment}: {Evidence} from a {Randomized} {Controlled} {Trial}}.{\BBCQ}
\newblock
\APACjournalVolNumPages{American Economic Review}{105}{2}{609--645}.
\newblock
\begin{APACrefURL} [{2023-10-10}]\url{https://pubs.aeaweb.org/doi/10.1257/aer.20130267} \end{APACrefURL}
\newblock
\begin{APACrefDOI} \doi{10.1257/aer.20130267} \end{APACrefDOI}
\PrintBackRefs{\CurrentBib}

\bibitem [\protect \citeauthoryear {%
K.~Cohen%
\ \BBA {} Leshem%
}{%
K.~Cohen%
\ \BBA {} Leshem%
}{%
{\protect \APACyear {2021}}%
}]{%
cohen_suppressing_2021}
\APACinsertmetastar {%
cohen_suppressing_2021}%
\begin{APACrefauthors}%
Cohen, K.%
\BCBT {}\ \BBA {} Leshem, A.%
\end{APACrefauthors}%
\unskip\
\newblock
\APACrefYearMonthDay{2021}{{\APACmonth{03}}}{}.
\newblock
{\BBOQ}\APACrefatitle {Suppressing the impact of the {COVID}-19 pandemic using controlled testing and isolation} {Suppressing the impact of the {COVID}-19 pandemic using controlled testing and isolation}.{\BBCQ}
\newblock
\APACjournalVolNumPages{Scientific Reports}{11}{1}{6279}.
\newblock
\begin{APACrefURL} \url{https://doi.org/10.1038/s41598-021-85458-1} \end{APACrefURL}
\newblock
\begin{APACrefDOI} \doi{10.1038/s41598-021-85458-1} \end{APACrefDOI}
\PrintBackRefs{\CurrentBib}

\bibitem [\protect \citeauthoryear {%
Dandl%
\ \protect \BOthers {.}}{%
Dandl%
\ \protect \BOthers {.}}{%
{\protect \APACyear {2022}}%
}]{%
dandl_what_2022}
\APACinsertmetastar {%
dandl_what_2022}%
\begin{APACrefauthors}%
Dandl, S.%
, Hothorn, T.%
, Seibold, H.%
, Sverdrup, E.%
, Wager, S.%
\BCBL {}\ \BBA {} Zeileis, A.%
\end{APACrefauthors}%
\unskip\
\newblock
\APACrefYearMonthDay{2022}{{\APACmonth{06}}}{}.
\newblock
\APACrefbtitle {What {Makes} {Forest}-{Based} {Heterogeneous} {Treatment} {Effect} {Estimators} {Work}?} {What {Makes} {Forest}-{Based} {Heterogeneous} {Treatment} {Effect} {Estimators} {Work}?}
\newblock
\APACaddressPublisher{}{arXiv}.
\newblock
\begin{APACrefURL} [{2023-02-17}]\url{http://arxiv.org/abs/2206.10323} \end{APACrefURL}
\newblock
\APACrefnote{arXiv:2206.10323 [stat]}
\PrintBackRefs{\CurrentBib}

\bibitem [\protect \citeauthoryear {%
Daulton%
, Balandat%
\BCBL {}\ \BBA {} Bakshy%
}{%
Daulton%
\ \protect \BOthers {.}}{%
{\protect \APACyear {2020}}%
}]{%
daulton_differentiable_2020}
\APACinsertmetastar {%
daulton_differentiable_2020}%
\begin{APACrefauthors}%
Daulton, S.%
, Balandat, M.%
\BCBL {}\ \BBA {} Bakshy, E.%
\end{APACrefauthors}%
\unskip\
\newblock
\APACrefYearMonthDay{2020}{}{}.
\newblock
{\BBOQ}\APACrefatitle {Differentiable {Expected} {Hypervolume} {Improvement} for {Parallel} {Multi}-{Objective} {Bayesian} {Optimization}} {Differentiable {Expected} {Hypervolume} {Improvement} for {Parallel} {Multi}-{Objective} {Bayesian} {Optimization}}.{\BBCQ}
\newblock
\BIn{} \APACrefbtitle {Proceedings of the 34th {International} {Conference} on {Neural} {Information} {Processing} {Systems}.} {Proceedings of the 34th {International} {Conference} on {Neural} {Information} {Processing} {Systems}.}
\newblock
\APACaddressPublisher{Red Hook, NY, USA}{Curran Associates Inc.}
\newblock
\APACrefnote{event-place: Vancouver, BC, Canada}
\PrintBackRefs{\CurrentBib}

\bibitem [\protect \citeauthoryear {%
Daulton%
, Balandat%
\BCBL {}\ \BBA {} Bakshy%
}{%
Daulton%
\ \protect \BOthers {.}}{%
{\protect \APACyear {2021}}%
}]{%
daulton_parallel_2021}
\APACinsertmetastar {%
daulton_parallel_2021}%
\begin{APACrefauthors}%
Daulton, S.%
, Balandat, M.%
\BCBL {}\ \BBA {} Bakshy, E.%
\end{APACrefauthors}%
\unskip\
\newblock
\APACrefYearMonthDay{2021}{{\APACmonth{10}}}{}.
\newblock
\APACrefbtitle {Parallel {Bayesian} {Optimization} of {Multiple} {Noisy} {Objectives} with {Expected} {Hypervolume} {Improvement}.} {Parallel {Bayesian} {Optimization} of {Multiple} {Noisy} {Objectives} with {Expected} {Hypervolume} {Improvement}.}
\newblock
\APACaddressPublisher{}{arXiv}.
\newblock
\begin{APACrefURL} [{2022-11-21}]\url{http://arxiv.org/abs/2105.08195} \end{APACrefURL}
\newblock
\APACrefnote{arXiv:2105.08195 [cs, stat]}
\PrintBackRefs{\CurrentBib}

\bibitem [\protect \citeauthoryear {%
Domingos%
}{%
Domingos%
}{%
{\protect \APACyear {1997}}%
}]{%
domingos_knowledge_1997}
\APACinsertmetastar {%
domingos_knowledge_1997}%
\begin{APACrefauthors}%
Domingos, P.%
\end{APACrefauthors}%
\unskip\
\newblock
\APACrefYearMonthDay{1997}{}{}.
\newblock
{\BBOQ}\APACrefatitle {Knowledge {Acquisition} {Form} {Examples} {Vis} {Multiple} {Models}} {Knowledge {Acquisition} {Form} {Examples} {Vis} {Multiple} {Models}}.{\BBCQ}
\newblock
\BIn{} \APACrefbtitle {Proceedings of the {Fourteenth} {International} {Conference} on {Machine} {Learning}} {Proceedings of the {Fourteenth} {International} {Conference} on {Machine} {Learning}}\ (\BPGS\ 98--106).
\newblock
\APACaddressPublisher{San Francisco, CA, USA}{Morgan Kaufmann Publishers Inc.}
\PrintBackRefs{\CurrentBib}

\bibitem [\protect \citeauthoryear {%
{Facebook}%
}{%
{Facebook}%
}{%
{\protect \APACyear {2022}}%
}]{%
facebook_ax_2022}
\APACinsertmetastar {%
facebook_ax_2022}%
\begin{APACrefauthors}%
{Facebook}.%
\end{APACrefauthors}%
\unskip\
\newblock
\APACrefYearMonthDay{2022}{}{}.
\newblock
\APACrefbtitle {Ax {Guide}.} {Ax {Guide}.}
\newblock
\begin{APACrefURL} [{2022-11-22}]\url{https://ax.dev//index.html} \end{APACrefURL}
\PrintBackRefs{\CurrentBib}

\bibitem [\protect \citeauthoryear {%
Fortnow%
}{%
Fortnow%
}{%
{\protect \APACyear {2013}}%
}]{%
fortnow_golden_2013}
\APACinsertmetastar {%
fortnow_golden_2013}%
\begin{APACrefauthors}%
Fortnow, L.%
\end{APACrefauthors}%
\unskip\
\newblock
\APACrefYear{2013}.
\newblock
\APACrefbtitle {The golden ticket: {P}, {NP}, and the search for the impossible} {The golden ticket: {P}, {NP}, and the search for the impossible}.
\newblock
\APACaddressPublisher{Princeton}{Princeton University Press}.
\PrintBackRefs{\CurrentBib}

\bibitem [\protect \citeauthoryear {%
Grubinger%
, Zeileis%
\BCBL {}\ \BBA {} Pfeiffer%
}{%
Grubinger%
\ \protect \BOthers {.}}{%
{\protect \APACyear {2014}}%
}]{%
grubinger_evtree_2014}
\APACinsertmetastar {%
grubinger_evtree_2014}%
\begin{APACrefauthors}%
Grubinger, T.%
, Zeileis, A.%
\BCBL {}\ \BBA {} Pfeiffer, K\BHBI P.%
\end{APACrefauthors}%
\unskip\
\newblock
\APACrefYearMonthDay{2014}{}{}.
\newblock
{\BBOQ}\APACrefatitle {evtree: {Evolutionary} {Learning} of {Globally} {Optimal} {Classification} and {Regression} {Trees} in {R}} {evtree: {Evolutionary} {Learning} of {Globally} {Optimal} {Classification} and {Regression} {Trees} in {R}}.{\BBCQ}
\newblock
\APACjournalVolNumPages{Journal of Statistical Software}{61}{1}{}.
\newblock
\begin{APACrefURL} [{2022-11-25}]\url{http://www.jstatsoft.org/v61/i01/} \end{APACrefURL}
\newblock
\begin{APACrefDOI} \doi{10.18637/jss.v061.i01} \end{APACrefDOI}
\PrintBackRefs{\CurrentBib}

\bibitem [\protect \citeauthoryear {%
Horn%
\ \BBA {} Bischl%
}{%
Horn%
\ \BBA {} Bischl%
}{%
{\protect \APACyear {2016}}%
}]{%
horn_multi-objective_2016}
\APACinsertmetastar {%
horn_multi-objective_2016}%
\begin{APACrefauthors}%
Horn, D.%
\BCBT {}\ \BBA {} Bischl, B.%
\end{APACrefauthors}%
\unskip\
\newblock
\APACrefYearMonthDay{2016}{{\APACmonth{12}}}{}.
\newblock
{\BBOQ}\APACrefatitle {Multi-objective parameter configuration of machine learning algorithms using model-based optimization} {Multi-objective parameter configuration of machine learning algorithms using model-based optimization}.{\BBCQ}
\newblock
\BIn{} \APACrefbtitle {2016 {IEEE} {Symposium} {Series} on {Computational} {Intelligence} ({SSCI})} {2016 {IEEE} {Symposium} {Series} on {Computational} {Intelligence} ({SSCI})}\ (\BPGS\ 1--8).
\newblock
\APACaddressPublisher{Athens, Greece}{IEEE}.
\newblock
\begin{APACrefURL} [{2022-11-15}]\url{http://ieeexplore.ieee.org/document/7850221/} \end{APACrefURL}
\newblock
\begin{APACrefDOI} \doi{10.1109/SSCI.2016.7850221} \end{APACrefDOI}
\PrintBackRefs{\CurrentBib}

\bibitem [\protect \citeauthoryear {%
Hu%
, Rudin%
\BCBL {}\ \BBA {} Seltzer%
}{%
Hu%
\ \protect \BOthers {.}}{%
{\protect \APACyear {2019}}%
}]{%
hu_optimal_2019}
\APACinsertmetastar {%
hu_optimal_2019}%
\begin{APACrefauthors}%
Hu, X.%
, Rudin, C.%
\BCBL {}\ \BBA {} Seltzer, M.%
\end{APACrefauthors}%
\unskip\
\newblock
\APACrefYearMonthDay{2019}{}{}.
\newblock
{\BBOQ}\APACrefatitle {Optimal sparse decision trees} {Optimal sparse decision trees}.{\BBCQ}
\newblock
\APACjournalVolNumPages{Advances in Neural Information Processing Systems (NeurIPS)}{}{}{}.
\PrintBackRefs{\CurrentBib}

\bibitem [\protect \citeauthoryear {%
Ireni~Saban%
\ \BBA {} Sherman%
}{%
Ireni~Saban%
\ \BBA {} Sherman%
}{%
{\protect \APACyear {2022}}%
}]{%
ireni_saban_ethical_2022}
\APACinsertmetastar {%
ireni_saban_ethical_2022}%
\begin{APACrefauthors}%
Ireni~Saban, L.%
\BCBT {}\ \BBA {} Sherman, M.%
\end{APACrefauthors}%
\unskip\
\newblock
\APACrefYear{2022}.
\newblock
\APACrefbtitle {Ethical governance of artificial intelligence in the public sector} {Ethical governance of artificial intelligence in the public sector}.
\newblock
\APACaddressPublisher{London ; New York}{Routledge, Taylor \& Francis Group}.
\PrintBackRefs{\CurrentBib}

\bibitem [\protect \citeauthoryear {%
Kallus%
}{%
Kallus%
}{%
{\protect \APACyear {2017}}%
}]{%
kallus_recursive_2017}
\APACinsertmetastar {%
kallus_recursive_2017}%
\begin{APACrefauthors}%
Kallus, N.%
\end{APACrefauthors}%
\unskip\
\newblock
\APACrefYearMonthDay{2017}{{\APACmonth{08}}}{}.
\newblock
\APACrefbtitle {Recursive {Partitioning} for {Personalization} using {Observational} {Data}.} {Recursive {Partitioning} for {Personalization} using {Observational} {Data}.}
\newblock
\APACaddressPublisher{}{arXiv}.
\newblock
\begin{APACrefURL} [{2022-11-21}]\url{http://arxiv.org/abs/1608.08925} \end{APACrefURL}
\newblock
\APACrefnote{arXiv:1608.08925 [cs, stat]}
\PrintBackRefs{\CurrentBib}

\bibitem [\protect \citeauthoryear {%
Karl%
\ \protect \BOthers {.}}{%
Karl%
\ \protect \BOthers {.}}{%
{\protect \APACyear {2022}}%
}]{%
karl_multi-objective_2022}
\APACinsertmetastar {%
karl_multi-objective_2022}%
\begin{APACrefauthors}%
Karl, F.%
, Pielok, T.%
, Moosbauer, J.%
, Pfisterer, F.%
, Coors, S.%
, Binder, M.%
\BDBL {}Bischl, B.%
\end{APACrefauthors}%
\unskip\
\newblock
\APACrefYearMonthDay{2022}{{\APACmonth{07}}}{}.
\newblock
\APACrefbtitle {Multi-{Objective} {Hyperparameter} {Optimization} -- {An} {Overview}.} {Multi-{Objective} {Hyperparameter} {Optimization} -- {An} {Overview}.}
\newblock
\APACaddressPublisher{}{arXiv}.
\newblock
\begin{APACrefURL} [{2022-11-21}]\url{http://arxiv.org/abs/2206.07438} \end{APACrefURL}
\newblock
\APACrefnote{arXiv:2206.07438 [cs, stat]}
\PrintBackRefs{\CurrentBib}

\bibitem [\protect \citeauthoryear {%
Kitagawa%
\ \BBA {} Tetenov%
}{%
Kitagawa%
\ \BBA {} Tetenov%
}{%
{\protect \APACyear {2018}}%
}]{%
kitagawa_who_2018}
\APACinsertmetastar {%
kitagawa_who_2018}%
\begin{APACrefauthors}%
Kitagawa, T.%
\BCBT {}\ \BBA {} Tetenov, A.%
\end{APACrefauthors}%
\unskip\
\newblock
\APACrefYearMonthDay{2018}{}{}.
\newblock
{\BBOQ}\APACrefatitle {Who {Should} {Be} {Treated}? {Empirical} {Welfare} {Maximization} {Methods} for {Treatment} {Choice}} {Who {Should} {Be} {Treated}? {Empirical} {Welfare} {Maximization} {Methods} for {Treatment} {Choice}}.{\BBCQ}
\newblock
\APACjournalVolNumPages{Econometrica}{86}{2}{591--616}.
\newblock
\begin{APACrefURL} [{2023-07-23}]\url{https://www.econometricsociety.org/doi/10.3982/ECTA13288} \end{APACrefURL}
\newblock
\begin{APACrefDOI} \doi{10.3982/ECTA13288} \end{APACrefDOI}
\PrintBackRefs{\CurrentBib}

\bibitem [\protect \citeauthoryear {%
Knaus%
, Lechner%
\BCBL {}\ \BBA {} Strittmatter%
}{%
Knaus%
\ \protect \BOthers {.}}{%
{\protect \APACyear {2021}}%
}]{%
knaus_machine_2021}
\APACinsertmetastar {%
knaus_machine_2021}%
\begin{APACrefauthors}%
Knaus, M\BPBI C.%
, Lechner, M.%
\BCBL {}\ \BBA {} Strittmatter, A.%
\end{APACrefauthors}%
\unskip\
\newblock
\APACrefYearMonthDay{2021}{{\APACmonth{03}}}{}.
\newblock
{\BBOQ}\APACrefatitle {Machine learning estimation of heterogeneous causal effects: {Empirical} {Monte} {Carlo} evidence} {Machine learning estimation of heterogeneous causal effects: {Empirical} {Monte} {Carlo} evidence}.{\BBCQ}
\newblock
\APACjournalVolNumPages{The Econometrics Journal}{24}{1}{134--161}.
\newblock
\begin{APACrefURL} [{2021-08-18}]\url{https://academic.oup.com/ectj/article/24/1/134/5854188} \end{APACrefURL}
\newblock
\begin{APACrefDOI} \doi{10.1093/ectj/utaa014} \end{APACrefDOI}
\PrintBackRefs{\CurrentBib}

\bibitem [\protect \citeauthoryear {%
Künzel%
, Sekhon%
, Bickel%
\BCBL {}\ \BBA {} Yu%
}{%
Künzel%
\ \protect \BOthers {.}}{%
{\protect \APACyear {2019}}%
}]{%
kunzel_metalearners_2019}
\APACinsertmetastar {%
kunzel_metalearners_2019}%
\begin{APACrefauthors}%
Künzel, S\BPBI R.%
, Sekhon, J\BPBI S.%
, Bickel, P\BPBI J.%
\BCBL {}\ \BBA {} Yu, B.%
\end{APACrefauthors}%
\unskip\
\newblock
\APACrefYearMonthDay{2019}{{\APACmonth{03}}}{}.
\newblock
{\BBOQ}\APACrefatitle {Metalearners for estimating heterogeneous treatment effects using machine learning} {Metalearners for estimating heterogeneous treatment effects using machine learning}.{\BBCQ}
\newblock
\APACjournalVolNumPages{Proceedings of the National Academy of Sciences}{116}{10}{4156--4165}.
\newblock
\begin{APACrefURL} [{2021-08-30}]\url{http://www.pnas.org/lookup/doi/10.1073/pnas.1804597116} \end{APACrefURL}
\newblock
\begin{APACrefDOI} \doi{10.1073/pnas.1804597116} \end{APACrefDOI}
\PrintBackRefs{\CurrentBib}

\bibitem [\protect \citeauthoryear {%
Lipton%
}{%
Lipton%
}{%
{\protect \APACyear {2018}}%
}]{%
lipton_mythos_2018}
\APACinsertmetastar {%
lipton_mythos_2018}%
\begin{APACrefauthors}%
Lipton, Z\BPBI C.%
\end{APACrefauthors}%
\unskip\
\newblock
\APACrefYearMonthDay{2018}{}{}.
\newblock
{\BBOQ}\APACrefatitle {The {Mythos} of {Model} {Interpretability}: {In} machine learning, the concept of interpretability is both important and slippery.} {The {Mythos} of {Model} {Interpretability}: {In} machine learning, the concept of interpretability is both important and slippery.}{\BBCQ}
\newblock
\APACjournalVolNumPages{Queue}{16}{3}{31--57}.
\newblock
\APACrefnote{Publisher: ACM New York, NY, USA}
\PrintBackRefs{\CurrentBib}

\bibitem [\protect \citeauthoryear {%
Liu%
\ \protect \BOthers {.}}{%
Liu%
\ \protect \BOthers {.}}{%
{\protect \APACyear {2014}}%
}]{%
liu_learning_2014}
\APACinsertmetastar {%
liu_learning_2014}%
\begin{APACrefauthors}%
Liu, S.%
, Dissanayake, S.%
, Patel, S.%
, Dang, X.%
, Mlsna, T.%
, Chen, Y.%
\BCBL {}\ \BBA {} Wilkins, D.%
\end{APACrefauthors}%
\unskip\
\newblock
\APACrefYearMonthDay{2014}{{\APACmonth{12}}}{}.
\newblock
{\BBOQ}\APACrefatitle {Learning accurate and interpretable models based on regularized random forests regression} {Learning accurate and interpretable models based on regularized random forests regression}.{\BBCQ}
\newblock
\APACjournalVolNumPages{BMC Systems Biology}{8}{S3}{S5}.
\newblock
\begin{APACrefURL} [{2023-02-01}]\url{https://bmcsystbiol.biomedcentral.com/articles/10.1186/1752-0509-8-S3-S5} \end{APACrefURL}
\newblock
\begin{APACrefDOI} \doi{10.1186/1752-0509-8-S3-S5} \end{APACrefDOI}
\PrintBackRefs{\CurrentBib}

\bibitem [\protect \citeauthoryear {%
Manski%
}{%
Manski%
}{%
{\protect \APACyear {2000}}%
}]{%
manski_identification_2000}
\APACinsertmetastar {%
manski_identification_2000}%
\begin{APACrefauthors}%
Manski, C\BPBI F.%
\end{APACrefauthors}%
\unskip\
\newblock
\APACrefYearMonthDay{2000}{{\APACmonth{04}}}{}.
\newblock
{\BBOQ}\APACrefatitle {Identification problems and decisions under ambiguity: {Empirical} analysis of treatment response and normative analysis of treatment choice} {Identification problems and decisions under ambiguity: {Empirical} analysis of treatment response and normative analysis of treatment choice}.{\BBCQ}
\newblock
\APACjournalVolNumPages{Journal of Econometrics}{95}{2}{415--442}.
\newblock
\begin{APACrefURL} [{2023-05-18}]\url{https://linkinghub.elsevier.com/retrieve/pii/S0304407699000457} \end{APACrefURL}
\newblock
\begin{APACrefDOI} \doi{10.1016/S0304-4076(99)00045-7} \end{APACrefDOI}
\PrintBackRefs{\CurrentBib}

\bibitem [\protect \citeauthoryear {%
Manski%
}{%
Manski%
}{%
{\protect \APACyear {2004}}%
}]{%
manski_statistical_2004}
\APACinsertmetastar {%
manski_statistical_2004}%
\begin{APACrefauthors}%
Manski, C\BPBI F.%
\end{APACrefauthors}%
\unskip\
\newblock
\APACrefYearMonthDay{2004}{{\APACmonth{07}}}{}.
\newblock
{\BBOQ}\APACrefatitle {Statistical {Treatment} {Rules} for {Heterogeneous} {Populations}} {Statistical {Treatment} {Rules} for {Heterogeneous} {Populations}}.{\BBCQ}
\newblock
\APACjournalVolNumPages{Econometrica}{72}{4}{1221--1246}.
\newblock
\begin{APACrefURL} [{2023-05-12}]\url{http://www.blackwell-synergy.com/links/doi/10.1111%2Fj.1468-0262.2004.00530.x} \end{APACrefURL}
\newblock
\begin{APACrefDOI} \doi{10.1111/j.1468-0262.2004.00530.x} \end{APACrefDOI}
\PrintBackRefs{\CurrentBib}

\bibitem [\protect \citeauthoryear {%
Manson%
, Chamberlain%
\BCBL {}\ \BBA {} Bourne%
}{%
Manson%
\ \protect \BOthers {.}}{%
{\protect \APACyear {2021}}%
}]{%
manson_mvmoo_2021}
\APACinsertmetastar {%
manson_mvmoo_2021}%
\begin{APACrefauthors}%
Manson, J\BPBI A.%
, Chamberlain, T\BPBI W.%
\BCBL {}\ \BBA {} Bourne, R\BPBI A.%
\end{APACrefauthors}%
\unskip\
\newblock
\APACrefYearMonthDay{2021}{{\APACmonth{08}}}{}.
\newblock
{\BBOQ}\APACrefatitle {{MVMOO}: {Mixed} variable multi-objective optimisation} {{MVMOO}: {Mixed} variable multi-objective optimisation}.{\BBCQ}
\newblock
\APACjournalVolNumPages{Journal of Global Optimization}{80}{4}{865--886}.
\newblock
\begin{APACrefURL} [{2022-11-25}]\url{https://link.springer.com/10.1007/s10898-021-01052-9} \end{APACrefURL}
\newblock
\begin{APACrefDOI} \doi{10.1007/s10898-021-01052-9} \end{APACrefDOI}
\PrintBackRefs{\CurrentBib}

\bibitem [\protect \citeauthoryear {%
Mehrabi%
, Morstatter%
, Saxena%
, Lerman%
\BCBL {}\ \BBA {} Galstyan%
}{%
Mehrabi%
\ \protect \BOthers {.}}{%
{\protect \APACyear {2019}}%
}]{%
mehrabi_survey_2019}
\APACinsertmetastar {%
mehrabi_survey_2019}%
\begin{APACrefauthors}%
Mehrabi, N.%
, Morstatter, F.%
, Saxena, N.%
, Lerman, K.%
\BCBL {}\ \BBA {} Galstyan, A.%
\end{APACrefauthors}%
\unskip\
\newblock
\APACrefYearMonthDay{2019}{}{}.
\newblock
{\BBOQ}\APACrefatitle {A survey on bias and fairness in machine learning} {A survey on bias and fairness in machine learning}.{\BBCQ}
\newblock
\APACjournalVolNumPages{arXiv preprint arXiv:1908.09635}{}{}{}.
\PrintBackRefs{\CurrentBib}

\bibitem [\protect \citeauthoryear {%
Mittelstadt%
, Allo%
, Taddeo%
, Wachter%
\BCBL {}\ \BBA {} Floridi%
}{%
Mittelstadt%
\ \protect \BOthers {.}}{%
{\protect \APACyear {2016}}%
}]{%
mittelstadt_ethics_2016}
\APACinsertmetastar {%
mittelstadt_ethics_2016}%
\begin{APACrefauthors}%
Mittelstadt, B\BPBI D.%
, Allo, P.%
, Taddeo, M.%
, Wachter, S.%
\BCBL {}\ \BBA {} Floridi, L.%
\end{APACrefauthors}%
\unskip\
\newblock
\APACrefYearMonthDay{2016}{{\APACmonth{12}}}{}.
\newblock
{\BBOQ}\APACrefatitle {The ethics of algorithms: {Mapping} the debate} {The ethics of algorithms: {Mapping} the debate}.{\BBCQ}
\newblock
\APACjournalVolNumPages{Big Data \& Society}{3}{2}{205395171667967}.
\newblock
\begin{APACrefURL} [{2022-12-15}]\url{http://journals.sagepub.com/doi/10.1177/2053951716679679} \end{APACrefURL}
\newblock
\begin{APACrefDOI} \doi{10.1177/2053951716679679} \end{APACrefDOI}
\PrintBackRefs{\CurrentBib}

\bibitem [\protect \citeauthoryear {%
Morales-Hernández%
, Van~Nieuwenhuyse%
\BCBL {}\ \BBA {} Gonzalez%
}{%
Morales-Hernández%
\ \protect \BOthers {.}}{%
{\protect \APACyear {2022}}%
}]{%
morales-hernandez_survey_2022}
\APACinsertmetastar {%
morales-hernandez_survey_2022}%
\begin{APACrefauthors}%
Morales-Hernández, A.%
, Van~Nieuwenhuyse, I.%
\BCBL {}\ \BBA {} Gonzalez, S\BPBI R.%
\end{APACrefauthors}%
\unskip\
\newblock
\APACrefYearMonthDay{2022}{{\APACmonth{11}}}{}.
\newblock
\APACrefbtitle {A survey on multi-objective hyperparameter optimization algorithms for {Machine} {Learning}.} {A survey on multi-objective hyperparameter optimization algorithms for {Machine} {Learning}.}
\newblock
\APACaddressPublisher{}{arXiv}.
\newblock
\begin{APACrefURL} [{2022-11-22}]\url{http://arxiv.org/abs/2111.13755} \end{APACrefURL}
\newblock
\APACrefnote{arXiv:2111.13755 [cs, math]}
\PrintBackRefs{\CurrentBib}

\bibitem [\protect \citeauthoryear {%
Neuman%
}{%
Neuman%
}{%
{\protect \APACyear {2014}}%
}]{%
neuman_social_2014}
\APACinsertmetastar {%
neuman_social_2014}%
\begin{APACrefauthors}%
Neuman, W\BPBI L.%
\end{APACrefauthors}%
\unskip\
\newblock
\APACrefYear{2014}.
\newblock
\APACrefbtitle {Social research methods: qualitative and quantitative approaches} {Social research methods: qualitative and quantitative approaches}\ (\PrintOrdinal{Pearson new international;Seventh;}\ \BEd)\ (\BNUMS\ Book, Whole).
\newblock
\APACaddressPublisher{Harlow, Essex}{Pearson}.
\PrintBackRefs{\CurrentBib}

\bibitem [\protect \citeauthoryear {%
Nie%
\ \BBA {} Wager%
}{%
Nie%
\ \BBA {} Wager%
}{%
{\protect \APACyear {2020}}%
}]{%
nie_quasi-oracle_2020}
\APACinsertmetastar {%
nie_quasi-oracle_2020}%
\begin{APACrefauthors}%
Nie, X.%
\BCBT {}\ \BBA {} Wager, S.%
\end{APACrefauthors}%
\unskip\
\newblock
\APACrefYearMonthDay{2020}{{\APACmonth{08}}}{}.
\newblock
{\BBOQ}\APACrefatitle {Quasi-{Oracle} {Estimation} of {Heterogeneous} {Treatment} {Effects}} {Quasi-{Oracle} {Estimation} of {Heterogeneous} {Treatment} {Effects}}.{\BBCQ}
\newblock
\APACjournalVolNumPages{arXiv:1712.04912 [econ, math, stat]}{}{}{}.
\newblock
\begin{APACrefURL} [{2021-05-17}]\url{http://arxiv.org/abs/1712.04912} \end{APACrefURL}
\newblock
\APACrefnote{arXiv: 1712.04912}
\PrintBackRefs{\CurrentBib}

\bibitem [\protect \citeauthoryear {%
Oprescu%
, Syrgkanis%
, Battocchi%
, Hei%
\BCBL {}\ \BBA {} Lewis%
}{%
Oprescu%
\ \protect \BOthers {.}}{%
{\protect \APACyear {2019}}%
}]{%
oprescu_econml_2019}
\APACinsertmetastar {%
oprescu_econml_2019}%
\begin{APACrefauthors}%
Oprescu, M.%
, Syrgkanis, V.%
, Battocchi, K.%
, Hei, M.%
\BCBL {}\ \BBA {} Lewis, G.%
\end{APACrefauthors}%
\unskip\
\newblock
\APACrefYearMonthDay{2019}{}{}.
\newblock
{\BBOQ}\APACrefatitle {{EconML}: {A} {Machine} {Learning} {Library} for {Estimating} {Heterogeneous} {Treatment} {Effects}} {{EconML}: {A} {Machine} {Learning} {Library} for {Estimating} {Heterogeneous} {Treatment} {Effects}}.{\BBCQ}
\newblock
\BIn{} \APACrefbtitle {33rd {Conference} on {Neural} {Information} {Processing} {Systems}} {33rd {Conference} on {Neural} {Information} {Processing} {Systems}}\ (\BPG~6).
\newblock
\APACaddressPublisher{Vancouver, Canada}{}.
\PrintBackRefs{\CurrentBib}

\bibitem [\protect \citeauthoryear {%
Plano~Clark%
\ \BBA {} Ivankova%
}{%
Plano~Clark%
\ \BBA {} Ivankova%
}{%
{\protect \APACyear {2022}}%
}]{%
plano_clark_mixed_2022}
\APACinsertmetastar {%
plano_clark_mixed_2022}%
\begin{APACrefauthors}%
Plano~Clark, V\BPBI L.%
\BCBT {}\ \BBA {} Ivankova, N\BPBI V.%
\end{APACrefauthors}%
\unskip\
\newblock
\APACrefYear{2022}.
\newblock
\APACrefbtitle {Mixed {Methods} {Research}: {A} {Guide} to the {Field}} {Mixed {Methods} {Research}: {A} {Guide} to the {Field}}.
\newblock
\APACaddressPublisher{Thousand Oaks, California}{}.
\newblock
\begin{APACrefURL} \url{https://methods.sagepub.com/book/mixed-methods-research-a-guide-to-the-field} \end{APACrefURL}
\newblock
\begin{APACrefDOI} \doi{10.4135/9781483398341} \end{APACrefDOI}
\PrintBackRefs{\CurrentBib}

\bibitem [\protect \citeauthoryear {%
Robins%
, Rotnitzky%
\BCBL {}\ \BBA {} Zhao%
}{%
Robins%
\ \protect \BOthers {.}}{%
{\protect \APACyear {1994}}%
}]{%
robins_estimation_1994}
\APACinsertmetastar {%
robins_estimation_1994}%
\begin{APACrefauthors}%
Robins, J\BPBI M.%
, Rotnitzky, A.%
\BCBL {}\ \BBA {} Zhao, L\BPBI P.%
\end{APACrefauthors}%
\unskip\
\newblock
\APACrefYearMonthDay{1994}{}{}.
\newblock
{\BBOQ}\APACrefatitle {Estimation of {Regression} {Coefficients} {When} {Some} {Regressors} {Are} {Not} {Always} {Observed}} {Estimation of {Regression} {Coefficients} {When} {Some} {Regressors} {Are} {Not} {Always} {Observed}}.{\BBCQ}
\newblock
\APACjournalVolNumPages{Journal of the American Statistical Association}{89}{427}{846--866}.
\newblock
\begin{APACrefURL} [{2021-07-20}]\url{http://www.jstor.org/stable/2290910} \end{APACrefURL}
\newblock
\APACrefnote{Publisher: [American Statistical Association, Taylor \& Francis, Ltd.]}
\newblock
\begin{APACrefDOI} \doi{10.2307/2290910} \end{APACrefDOI}
\PrintBackRefs{\CurrentBib}

\bibitem [\protect \citeauthoryear {%
Rudin%
}{%
Rudin%
}{%
{\protect \APACyear {2019}}%
}]{%
rudin_stop_2019}
\APACinsertmetastar {%
rudin_stop_2019}%
\begin{APACrefauthors}%
Rudin, C.%
\end{APACrefauthors}%
\unskip\
\newblock
\APACrefYearMonthDay{2019}{{\APACmonth{05}}}{}.
\newblock
{\BBOQ}\APACrefatitle {Stop explaining black box machine learning models for high stakes decisions and use interpretable models instead} {Stop explaining black box machine learning models for high stakes decisions and use interpretable models instead}.{\BBCQ}
\newblock
\APACjournalVolNumPages{Nature Machine Intelligence}{1}{5}{206--215}.
\newblock
\begin{APACrefURL} \url{https://doi.org/10.1038/s42256-019-0048-x} \end{APACrefURL}
\newblock
\begin{APACrefDOI} \doi{10.1038/s42256-019-0048-x} \end{APACrefDOI}
\PrintBackRefs{\CurrentBib}

\bibitem [\protect \citeauthoryear {%
Simon%
}{%
Simon%
}{%
{\protect \APACyear {1956}}%
}]{%
simon_rational_1956}
\APACinsertmetastar {%
simon_rational_1956}%
\begin{APACrefauthors}%
Simon, H\BPBI A.%
\end{APACrefauthors}%
\unskip\
\newblock
\APACrefYearMonthDay{1956}{}{}.
\newblock
{\BBOQ}\APACrefatitle {Rational choice and the structure of the environment.} {Rational choice and the structure of the environment.}{\BBCQ}
\newblock
\APACjournalVolNumPages{Psychological Review}{63}{2}{129--138}.
\newblock
\begin{APACrefURL} [{2023-07-23}]\url{http://doi.apa.org/getdoi.cfm?doi=10.1037/h0042769} \end{APACrefURL}
\newblock
\begin{APACrefDOI} \doi{10.1037/h0042769} \end{APACrefDOI}
\PrintBackRefs{\CurrentBib}

\bibitem [\protect \citeauthoryear {%
Su%
, Mou%
, Ding%
\BCBL {}\ \BBA {} Wainwright%
}{%
Su%
\ \protect \BOthers {.}}{%
{\protect \APACyear {2023}}%
}]{%
su_when_2023}
\APACinsertmetastar {%
su_when_2023}%
\begin{APACrefauthors}%
Su, F.%
, Mou, W.%
, Ding, P.%
\BCBL {}\ \BBA {} Wainwright, M\BPBI J.%
\end{APACrefauthors}%
\unskip\
\newblock
\APACrefYearMonthDay{2023}{{\APACmonth{03}}}{}.
\newblock
\APACrefbtitle {When is the estimated propensity score better? {High}-dimensional analysis and bias correction.} {When is the estimated propensity score better? {High}-dimensional analysis and bias correction.}
\newblock
\APACaddressPublisher{}{arXiv}.
\newblock
\begin{APACrefURL} [{2023-10-14}]\url{http://arxiv.org/abs/2303.17102} \end{APACrefURL}
\newblock
\APACrefnote{arXiv:2303.17102 [stat]}
\PrintBackRefs{\CurrentBib}

\bibitem [\protect \citeauthoryear {%
Sverdrup%
, Kanodia%
, Zhou%
, Athey%
\BCBL {}\ \BBA {} Wager%
}{%
Sverdrup%
\ \protect \BOthers {.}}{%
{\protect \APACyear {2020}}%
}]{%
sverdrup_policytree_2020}
\APACinsertmetastar {%
sverdrup_policytree_2020}%
\begin{APACrefauthors}%
Sverdrup, E.%
, Kanodia, A.%
, Zhou, Z.%
, Athey, S.%
\BCBL {}\ \BBA {} Wager, S.%
\end{APACrefauthors}%
\unskip\
\newblock
\APACrefYearMonthDay{2020}{{\APACmonth{06}}}{}.
\newblock
{\BBOQ}\APACrefatitle {policytree: {Policy} learning via doubly robust empirical welfare maximization over trees} {policytree: {Policy} learning via doubly robust empirical welfare maximization over trees}.{\BBCQ}
\newblock
\APACjournalVolNumPages{Journal of Open Source Software}{5}{50}{2232}.
\newblock
\begin{APACrefURL} [{2022-11-14}]\url{https://joss.theoj.org/papers/10.21105/joss.02232} \end{APACrefURL}
\newblock
\begin{APACrefDOI} \doi{10.21105/joss.02232} \end{APACrefDOI}
\PrintBackRefs{\CurrentBib}

\bibitem [\protect \citeauthoryear {%
Verhaeghe%
, Nijssen%
, Pesant%
, Quimper%
\BCBL {}\ \BBA {} Schaus%
}{%
Verhaeghe%
\ \protect \BOthers {.}}{%
{\protect \APACyear {2020}}%
}]{%
verhaeghe_learning_2020}
\APACinsertmetastar {%
verhaeghe_learning_2020}%
\begin{APACrefauthors}%
Verhaeghe, H.%
, Nijssen, S.%
, Pesant, G.%
, Quimper, C\BHBI G.%
\BCBL {}\ \BBA {} Schaus, P.%
\end{APACrefauthors}%
\unskip\
\newblock
\APACrefYearMonthDay{2020}{{\APACmonth{12}}}{}.
\newblock
{\BBOQ}\APACrefatitle {Learning optimal decision trees using constraint programming} {Learning optimal decision trees using constraint programming}.{\BBCQ}
\newblock
\APACjournalVolNumPages{Constraints}{25}{3-4}{226--250}.
\newblock
\begin{APACrefURL} [{2022-11-14}]\url{http://link.springer.com/10.1007/s10601-020-09312-3} \end{APACrefURL}
\newblock
\begin{APACrefDOI} \doi{10.1007/s10601-020-09312-3} \end{APACrefDOI}
\PrintBackRefs{\CurrentBib}

\bibitem [\protect \citeauthoryear {%
Wager%
\ \BBA {} Athey%
}{%
Wager%
\ \BBA {} Athey%
}{%
{\protect \APACyear {2018}}%
}]{%
wager_estimation_2018}
\APACinsertmetastar {%
wager_estimation_2018}%
\begin{APACrefauthors}%
Wager, S.%
\BCBT {}\ \BBA {} Athey, S.%
\end{APACrefauthors}%
\unskip\
\newblock
\APACrefYearMonthDay{2018}{{\APACmonth{07}}}{}.
\newblock
{\BBOQ}\APACrefatitle {Estimation and {Inference} of {Heterogeneous} {Treatment} {Effects} using {Random} {Forests}} {Estimation and {Inference} of {Heterogeneous} {Treatment} {Effects} using {Random} {Forests}}.{\BBCQ}
\newblock
\APACjournalVolNumPages{Journal of the American Statistical Association}{113}{523}{1228--1242}.
\newblock
\begin{APACrefURL} \url{https://doi.org/10.1080/01621459.2017.1319839} \end{APACrefURL}
\newblock
\APACrefnote{Publisher: Taylor \& Francis}
\newblock
\begin{APACrefDOI} \doi{10.1080/01621459.2017.1319839} \end{APACrefDOI}
\PrintBackRefs{\CurrentBib}

\bibitem [\protect \citeauthoryear {%
Wald%
}{%
Wald%
}{%
{\protect \APACyear {1949}}%
}]{%
wald_statistical_1949}
\APACinsertmetastar {%
wald_statistical_1949}%
\begin{APACrefauthors}%
Wald, A.%
\end{APACrefauthors}%
\unskip\
\newblock
\APACrefYearMonthDay{1949}{{\APACmonth{06}}}{}.
\newblock
{\BBOQ}\APACrefatitle {Statistical {Decision} {Functions}} {Statistical {Decision} {Functions}}.{\BBCQ}
\newblock
\APACjournalVolNumPages{The Annals of Mathematical Statistics}{20}{2}{165--205}.
\newblock
\begin{APACrefURL} [{2023-05-18}]\url{http://projecteuclid.org/euclid.aoms/1177730030} \end{APACrefURL}
\newblock
\begin{APACrefDOI} \doi{10.1214/aoms/1177730030} \end{APACrefDOI}
\PrintBackRefs{\CurrentBib}

\bibitem [\protect \citeauthoryear {%
Westreich%
, Lessler%
\BCBL {}\ \BBA {} Funk%
}{%
Westreich%
\ \protect \BOthers {.}}{%
{\protect \APACyear {2010}}%
}]{%
westreich_propensity_2010}
\APACinsertmetastar {%
westreich_propensity_2010}%
\begin{APACrefauthors}%
Westreich, D.%
, Lessler, J.%
\BCBL {}\ \BBA {} Funk, M\BPBI J.%
\end{APACrefauthors}%
\unskip\
\newblock
\APACrefYearMonthDay{2010}{}{}.
\newblock
{\BBOQ}\APACrefatitle {Propensity score estimation: machine learning and classification methods as alternatives to logistic regression} {Propensity score estimation: machine learning and classification methods as alternatives to logistic regression}.{\BBCQ}
\newblock
\APACjournalVolNumPages{Journal of clinical epidemiology}{63}{8}{826--833}.
\newblock
\begin{APACrefDOI} \doi{10.1016/j.jclinepi.2009.11.020} \end{APACrefDOI}
\PrintBackRefs{\CurrentBib}

\bibitem [\protect \citeauthoryear {%
A.~Zhou%
\ \protect \BOthers {.}}{%
A.~Zhou%
\ \protect \BOthers {.}}{%
{\protect \APACyear {2011}}%
}]{%
zhou_multiobjective_2011}
\APACinsertmetastar {%
zhou_multiobjective_2011}%
\begin{APACrefauthors}%
Zhou, A.%
, Qu, B\BHBI Y.%
, Li, H.%
, Zhao, S\BHBI Z.%
, Suganthan, P\BPBI N.%
\BCBL {}\ \BBA {} Zhang, Q.%
\end{APACrefauthors}%
\unskip\
\newblock
\APACrefYearMonthDay{2011}{{\APACmonth{03}}}{}.
\newblock
{\BBOQ}\APACrefatitle {Multiobjective evolutionary algorithms: {A} survey of the state of the art} {Multiobjective evolutionary algorithms: {A} survey of the state of the art}.{\BBCQ}
\newblock
\APACjournalVolNumPages{Swarm and Evolutionary Computation}{1}{1}{32--49}.
\newblock
\begin{APACrefURL} [{2022-11-21}]\url{https://linkinghub.elsevier.com/retrieve/pii/S2210650211000058} \end{APACrefURL}
\newblock
\begin{APACrefDOI} \doi{10.1016/j.swevo.2011.03.001} \end{APACrefDOI}
\PrintBackRefs{\CurrentBib}

\bibitem [\protect \citeauthoryear {%
Z.~Zhou%
, Athey%
\BCBL {}\ \BBA {} Wager%
}{%
Z.~Zhou%
\ \protect \BOthers {.}}{%
{\protect \APACyear {2023}}%
}]{%
zhou_offline_2023}
\APACinsertmetastar {%
zhou_offline_2023}%
\begin{APACrefauthors}%
Zhou, Z.%
, Athey, S.%
\BCBL {}\ \BBA {} Wager, S.%
\end{APACrefauthors}%
\unskip\
\newblock
\APACrefYearMonthDay{2023}{{\APACmonth{01}}}{}.
\newblock
{\BBOQ}\APACrefatitle {Offline {Multi}-{Action} {Policy} {Learning}: {Generalization} and {Optimization}} {Offline {Multi}-{Action} {Policy} {Learning}: {Generalization} and {Optimization}}.{\BBCQ}
\newblock
\APACjournalVolNumPages{Operations Research}{71}{1}{148--183}.
\newblock
\begin{APACrefURL} [{2023-08-10}]\url{https://pubsonline.informs.org/doi/10.1287/opre.2022.2271} \end{APACrefURL}
\newblock
\begin{APACrefDOI} \doi{10.1287/opre.2022.2271} \end{APACrefDOI}
\PrintBackRefs{\CurrentBib}

\bibitem [\protect \citeauthoryear {%
Zivich%
\ \BBA {} Breskin%
}{%
Zivich%
\ \BBA {} Breskin%
}{%
{\protect \APACyear {2021}}%
}]{%
zivich_machine_2021}
\APACinsertmetastar {%
zivich_machine_2021}%
\begin{APACrefauthors}%
Zivich, P\BPBI N.%
\BCBT {}\ \BBA {} Breskin, A.%
\end{APACrefauthors}%
\unskip\
\newblock
\APACrefYearMonthDay{2021}{{\APACmonth{05}}}{}.
\newblock
{\BBOQ}\APACrefatitle {Machine learning for causal inference: on the use of cross-fit estimators} {Machine learning for causal inference: on the use of cross-fit estimators}.{\BBCQ}
\newblock
\APACjournalVolNumPages{Epidemiology}{32}{3}{393--401}.
\newblock
\begin{APACrefURL} [{2021-08-25}]\url{http://arxiv.org/abs/2004.10337} \end{APACrefURL}
\newblock
\APACrefnote{arXiv: 2004.10337}
\newblock
\begin{APACrefDOI} \doi{10.1097/EDE.0000000000001332} \end{APACrefDOI}
\PrintBackRefs{\CurrentBib}

\bibitem [\protect \citeauthoryear {%
Zuhal%
, Palar%
\BCBL {}\ \BBA {} Shimoyama%
}{%
Zuhal%
\ \protect \BOthers {.}}{%
{\protect \APACyear {2019}}%
}]{%
zuhal_comparative_2019}
\APACinsertmetastar {%
zuhal_comparative_2019}%
\begin{APACrefauthors}%
Zuhal, L\BPBI R.%
, Palar, P\BPBI S.%
\BCBL {}\ \BBA {} Shimoyama, K.%
\end{APACrefauthors}%
\unskip\
\newblock
\APACrefYearMonthDay{2019}{{\APACmonth{08}}}{}.
\newblock
{\BBOQ}\APACrefatitle {A comparative study of multi-objective expected improvement for aerodynamic design} {A comparative study of multi-objective expected improvement for aerodynamic design}.{\BBCQ}
\newblock
\APACjournalVolNumPages{Aerospace Science and Technology}{91}{}{548--560}.
\newblock
\begin{APACrefURL} [{2022-11-22}]\url{https://linkinghub.elsevier.com/retrieve/pii/S1270963818327809} \end{APACrefURL}
\newblock
\begin{APACrefDOI} \doi{10.1016/j.ast.2019.05.044} \end{APACrefDOI}
\PrintBackRefs{\CurrentBib}

\end{thebibliography}

\newpage\section*{Appendix A - Variables used in application} \label{appendix}

\begin{table}[ht]
\centering
\begin{tabular}{rllll}
  \hline
 Name & Label & All & Selected & \shortstack{Just age\\and distance\\from clinic} \\ 
  \hline
head\_fem & hh head is female & X & X &  \\ 
  B\_head\_age & age of head & X &  &  \\ 
  B\_head\_age\_imputed & age of head with missing replaced by sample mean & X & X &  \\ 
  B\_head\_age\_missing & age of head missing & X &  &  \\ 
  headage\_x & Deciles of HH head age (11=missing). Includes all tmts & X &  &  \\ 
  head\_edu & years of education of household head & X &  &  \\ 
  head\_lit & head can read and write & X &  &  \\ 
  B\_knowledge\_correct & knows only mosquitoes transmit malaria & X &  &  \\ 
  head\_mar & interviewed head married & X & X &  \\ 
  head\_dep & number of dependents & X & X &  \\ 
  subfarm & main occupation is subsistence farming & X & X &  \\ 
  B\_hh\_size & \# of hh members & X & X &  \\ 
  B\_adultteen & number of adult and teen members (age$>$=9)  in hh & X & X &  \\ 
  head\_acres & acres of land & X & X &  \\ 
  B\_dist\_km & distance (km) from household to study chemist & X &  &  \\ 
  dist\_clinic & distance to closest health center & X & X & X \\ 
  num\_bednets & number of bednets owned & X &  &  \\ 
  share\_undernet & baseline: share of hh members sleeping under net (computed) & X &  &  \\ 
  B\_heard\_act & heard of ACTs & X &  &  \\ 
  B\_act\_best & from baseline A26: would buy ACT if money wasn't problem & X &  &  \\ 
  B\_heard\_rdt & heard of RDTs & X &  &  \\ 
  base\_mtest & someone took either RDT or microscopy test in past month & X &  &  \\ 
  base\_rdt & someone in household took RDT in past month & X &  &  \\ 
  base\_micro & someone in household took microscopy test in past month & X &  &  \\ 
  treat\_h2o & baseline a14: treated drinking water in past month & X &  &  \\ 
  B\_mal\_ct & baseline table C: total malaria episodes in hh & X &  &  \\ 
  B\_antimal\_cost & average cost across malaria episodes for which cost provided & X & &  \\ 
  B\_mal\_episode & baseline table C: any malaria episode in hh & X & &  \\ 
  B\_mal\_baby\_hh & number of babies (age$<$4) who had malaria & X &  &  \\ 
  B\_mal\_kid\_hh & number of kids (age$<$9 \& age$>$=4) who had malaria & X &  &  \\ 
  B\_mal\_teen\_hh & number of teen members (age$<$14 \& age$>$=9) who had malaria & X &X &  \\ 
  B\_mal\_adult\_hh & number of adult members (age$>$14) who had malaria & X &  &  \\ 
  LOG\_patient\_age1 & admin data: age of patient, 1st log entry for the hh & X & X & X \\ 
   \hline
\end{tabular}
\caption{\label{tab:appendix}Variables used in application (all variables measured at baseline except patient age).}
\end{table}

\end{document}